\documentclass[10pt,twocolumn,letterpaper]{article}
\usepackage{amsmath}   
\usepackage{amssymb}   
\usepackage{amsfonts}  
\usepackage{subcaption}
\usepackage{cvpr}              
\definecolor{cvprblue}{rgb}{0.21,0.49,0.74}
\usepackage[pagebackref,breaklinks,colorlinks,allcolors=cvprblue]{hyperref}

\def\confName{CVPR}
\def\confYear{2026}

\title{Beyond Voxel 3D Editing : Learning from 3D Masks and Self-Constructed Data}
\author{
    Yizhao Xu$^{1,3\star}$ \quad 
    Hongyuan Zhu$^{3\dagger}$ \quad 
    Caiyun Liu$^1$ \quad 
    Tianfu Wang$^2$ \quad 
    Keyu Chen$^1$ \\
    Sicheng Xu$^4$ \quad 
    Jiaolong Yang$^4$ \quad 
    Nicholas Jing Yuan$^3$ \quad 
    Qi Zhang$^3$ \\[0.5em]
    $^1$The Peking University \hfill
    $^2$ HKUST(GZ) \hfill
    $^3$Microsoft AI \hfill
    $^4$Microsoft Research     
}

\begin{document}


\twocolumn[{%
\renewcommand\twocolumn[1][]{#1}%
\maketitle
\begin{center}
    \includegraphics[width=\textwidth]{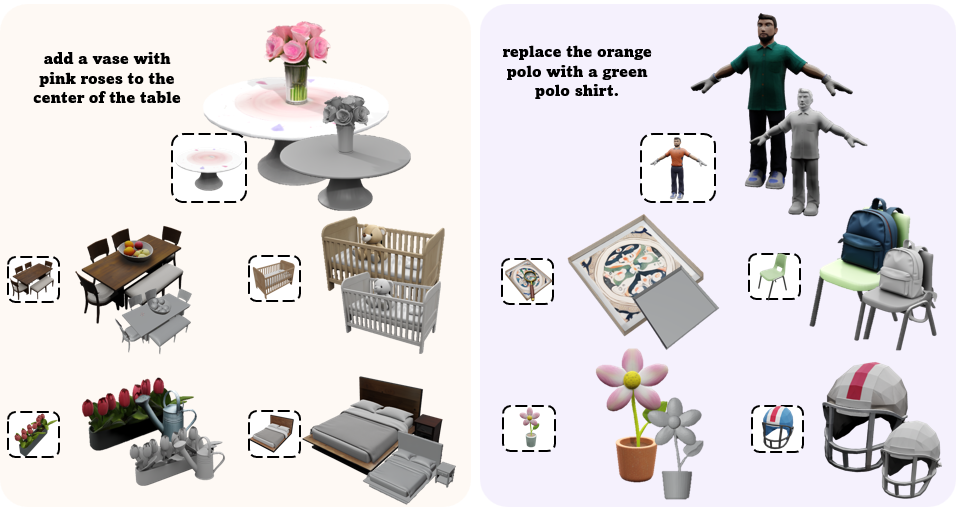}
    \includegraphics[width=\textwidth]{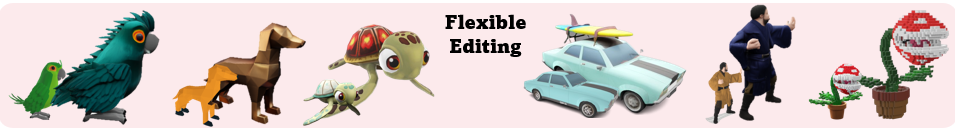}
    \captionof{figure}{Fast and versatile 3D editing results. Our method supports both global (e.g., add operation, left) and local (e.g., replacement or style change operation, right) edits in seconds. It yields high-quality assets with detailed meshes for geometry and vivid 3D Gaussians/Radiance Fields for appearance.}
    \label{fig:placeholder}
\end{center}
}]
\def\thefootnote{$\star$}\footnotetext{Work done during internship at Microsoft AI.}
\def\thefootnote{$\dagger$}\footnotetext{Corresponding author.}
\def\thefootnote{\arabic{footnote}} 

\begin{abstract}
3D editing refers to the ability to apply local or global modifications to 3D assets. Effective 3D editing requires maintaining semantic consistency by performing localized changes according to prompts, while also preserving local invariance so that unchanged regions remain consistent with the original. However, existing approaches have significant limitations: multi-view editing methods incur losses when projecting back to 3D, while voxel-based editing is constrained in both the regions that can be modified and the scale of modifications. Moreover, the lack of sufficiently large editing datasets for training and evaluation remains a challenge. To address these challenges, we propose a Beyond Voxel 3D Editing (BVE) framework with a self-constructed large-scale dataset specifically tailored for 3D editing. Building upon this dataset, our model enhances a foundational image-to-3D generative architecture with lightweight, trainable modules, enabling efficient injection of textual semantics without the need for expensive full-model retraining. Furthermore, we introduce an annotation-free 3D masking strategy to preserve local invariance, maintaining the integrity of unchanged regions during editing. Extensive experiments demonstrate that BVE achieves superior performance in generating high-quality, text-aligned 3D assets, while faithfully retaining the visual characteristics of the original input.

\end{abstract}    
\section{Introduction}
\label{sec:intro}
Generative artificial intelligence is revolutionizing 3D content creation, fundamentally reshaping workflows across various domains. 3D generation is widely applied across creative domains such as 3D printing~\cite{son20243d}, gaming~\cite{lai2025hunyuan3d}, virtual reality (VR), and augmented reality (AR). Recently, a surge of research~\cite{hao2024meshtron,chen2024meshanything,huang2024epidiff,xiang2024structured,zhao2025hunyuan3d, li2025step1x} and development in 3D generation has made it possible to produce high-resolution 3D assets with physically based rendering (PBR)~\cite{walter2007microfacet,he2025neural,xiong2025texgaussian,chen20253dtopia} materials.

Users typically create 3D assets from image or text prompts using generative tools. However, it is still challenging to generate high-quality assets in a single attempt — the results often require refinement in professional modeling software such as Blender or 3Ds Max to meet production standards. To better address user needs and enhance the flexibility of 3D generation pipelines, recent studies have begun exploring 3D editing capabilities, enabling more efficient and controllable asset creation workflows.

Current text-driven 3D editing approaches primarily fall into two categories, each with inherent trade-offs. The first, leveraging per-instance optimization like Score Distillation Sampling (SDS)~\cite{poole2022dreamfusion,chen2024shap,dong2024interactive3d}, iteratively refines a 3D asset to match the target prompt. While effective, this process is computationally prohibitive, rendering it impractical for interactive use. To enhance efficiency, a second category of methods pursues an indirect strategy: editing 2D rendered views and then lifting these changes back into 3D space~\cite{bar2025editp23, chen2024dge, chen2024generic}. However, this approach often struggles with cross-view consistency, leading to geometric artifacts and textural inconsistencies when fusing the edits into a coherent 3D model~\cite{liu2024sketchdream,mikaeili2023sked,chen2025partgen}. More recently, direct manipulation of 3D-native representations, such as voxel-level features in VoxHammer~\cite{li2025voxhammer}, has emerged as a promising direction. Yet, these methods still face the critical challenge of achieving precise local edits while preserving the global structure and semantic integrity of unchanged regions.

Indeed, 3D editing still faces significant challenges:
\begin{itemize}
    \item \textbf{Limited datasets:} There is a lack of sufficient 3D editing datasets for both training and evaluation.
    \item \textbf{Semantic consistency:} Effective 3D editing requires making localized changes according to prompts while preserving the semantic and structural integrity of unchanged regions.
    \item \textbf{Flexible editing:} Methods need to support both global and local editing of 3D assets.
\end{itemize}

To this end, we propose Beyond Voxel 3D Editing (BVE), an efficient framework that achieves high-quality, text-driven 3D editing while faithfully preserving the original asset's identity. Our approach integrates lightweight modules into a pretrained generative model and is validated on a large-scale benchmark we developed specifically for this task. In summary, our contributions are as follows:
\begin{itemize}
\item We construct and will release Edit-3DVerse, the first large-scale, high-quality dataset for text-driven 3D editing, purpose-built to benchmark research in this domain.
\item We propose the BVE framework, which enables high-fidelity 3D editing via lightweight, trainable modules, eliminating the need for costly full-model retraining.
\item We design a novel annotation-free 3D masking strategy that ensures semantic and structural consistency by precisely preserving unedited regions.
\item We demonstrate through extensive experiments that our method significantly outperforms state-of-the-art approaches in both editing quality and identity preservation.
\end{itemize}

\begin{figure*}[t]
  \centering
    \centering
    \includegraphics[width=\textwidth]{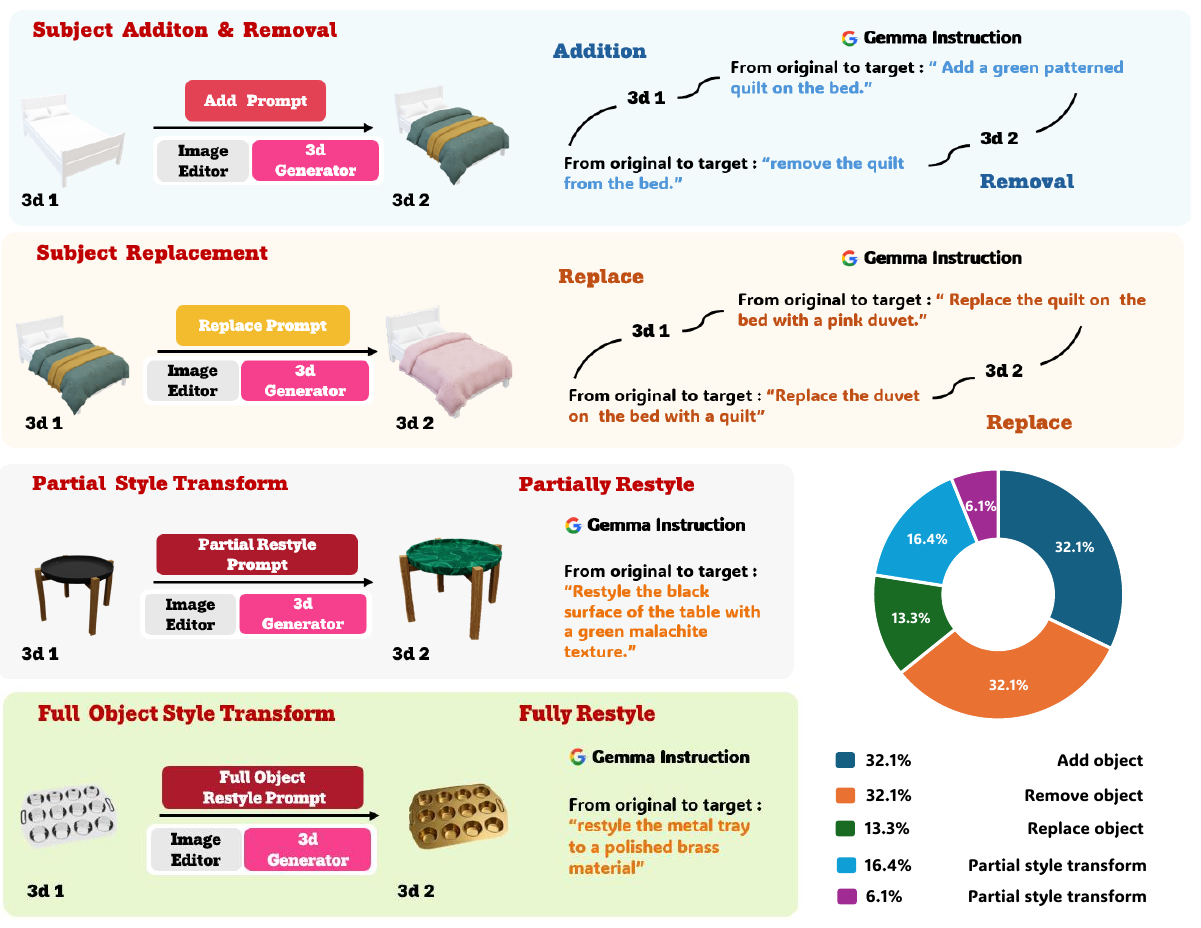}
    \caption{Dataset Construction pipeline and Sub-Task Distribution}
\end{figure*}

\section{Related Work}

\subsection{3D Generative Models}
Recent advancements in diffusion models \cite{ho2020denoising,song2022denoisingdiffusionimplicitmodels} and the availability of high-quality 3D datasets \cite{deitke2022objaverseuniverseannotated3d,deitke2023objaverse} have significantly accelerated the development of 3D generative modeling \cite{chen2024meshxlneuralcoordinatefield, chen2024generic, dong2024telatextlayerwise3d, gao2025meshartgeneratingarticulatedmeshes, hao2024meshtronhighfidelityartistlike3d, hong2024lrmlargereconstructionmodel, huang2024epidiffenhancingmultiviewsynthesis, li2025craftsman3dhighfidelitymeshgeneration,li2025step1x,liu2023one2345fastsingleimage,liu2023one2345singleimage3d,liu2024syncdreamergeneratingmultiviewconsistentimages, long2023wonder3dsingleimage3d,meng2025lt3sdlatenttrees3d,Roessle_2024,tang2024lgmlargemultiviewgaussian,voleti2024sv3dnovelmultiviewsynthesis, wang2024llamameshunifying3dmesh,wang2024crmsingleimage3d,wei2025octgptoctreebasedmultiscaleautoregressive,wu2024unique3dhighqualityefficient3d,wu2025dipodualstateimagescontrolled,wu2024direct3dscalableimageto3dgeneration,wu2024blockfusionexpandable3dscene,xu2024instantmeshefficient3dmesh,ye2025shapellmomninativemultimodalllm,zhang2024claycontrollablelargescalegenerative,zhao2025deepmeshautoregressiveartistmeshcreation}. These methods can generally be categorized into two major paradigms. One approach involves generating 3D models by first synthesizing multi-view images and then reconstructing the 3D from these views \cite{huang2025stereogsmultiviewstereovision,huang2024mvadaptermultiviewconsistentimage,liu2024syncdreamergeneratingmultiviewconsistentimages,long2023wonder3dsingleimage3d,qu2025deocc1to33ddeocclusionsingle,radford2021learningtransferablevisualmodels}. However, inconsistent multi-view synthesis may lower the quality of the final 3D model. Alternatively, a series of methods focuses on training native 3D generative models \cite{li2025craftsman3dhighfidelitymeshgeneration,li2025triposg,li2025triposghighfidelity3dshape,lin2025partcrafterstructured3dmesh,wu2025dipodualstateimagescontrolled,wu2024direct3dscalableimageto3dgeneration,wu2025direct3ds2gigascale3dgeneration, zhao2025assemblerscalable3dassembly,zhao2023michelangeloconditional3dshape}. These often comprise a variational autoencoder \cite{kingma2022autoencodingvariationalbayes} and a diffusion transformer (DiT) \cite{peebles2023scalablediffusionmodelstransformers} for denoising in latent space. This approach effectively unifies 3D generation with high fidelity and consistency, laying the foundation for downstream inversion and editing.

\subsection{3D Editing}
Compared to traditional image editing tasks~\cite{zhao2024ultraedit,ju2024brushnet,shi2024seededit}, 3D editing remains relatively underexplored. This is partly due to the limited availability of high-quality base models and 3D-related datasets. Recently, with advancements in 3D generation models, several works on 3D editing have begun to emerge. For example, work based on Score Distillation Sampling (SDS)\cite{poole2022dreamfusion} often requires substantial computational resources to achieve effective editing. In contrast, more recent research~\cite{bar2025editp23, chen2024dge, chen2024generic,barda2024instant3ditmultiviewinpaintingfast,cao2024mvinpainterlearningmultiviewconsistent,erkoç2024preditor3dfastprecise3d,li2025cmdcontrollablemultiviewdiffusion,zheng2025pro3deditorprogressiveviewsperspective} explores multi-view image editing as an indirect approach to modifying 3D assets. This strategy edits 3D content by altering its associated images, but it can introduce noise due to inconsistencies when projecting multiple views back onto the 3D asset. Recently, TRELLIS proposes a two-stage framework using SS and Slat structures for 3D asset generation. Additionally, several training-free methods~\cite{ye2025nano3d,li2025voxhammer} have been introduced that edit voxel-level features directly. For instance, Voxhammer~\cite{li2025voxhammer} merges the inverted sampling of the unmasked regions with the generated masked regions to achieve controlled 3D asset edits.

\section{Method}
\label{sec:method}
\subsection{Data Creation}
\label{sec:data_creation}

To address limitations in existing 3D editing datasets, we developed an automated data generation pipeline that yielded Edit3D-Verse, a diverse dataset with over 100k high-quality samples curated from an initial pool of over 500k. Our pipeline ensures quality through a three-stage framework: (1) Prompt Construction, (2) Image Generation, and (3) 3D Generation, as illustrated in \cref{fig:pipeline}.

\subsubsection*{Prompt Construction.}
We first render a source 3D asset into multi-view images, which are filtered by a VLM to discard low-quality views. For the remaining views, a dual-branch framework generates instructions. One branch uses Gemma 3\cite{gemma_2025} for global edits like \textit{Add} or \textit{Restyle}, while the other targets localized edits by employing SAM-2\cite{ravi2024sam2} to isolate object parts and Florence-2\cite{xiao2023florence} to caption them, facilitating precise \textit{Replace} prompts.

\subsubsection*{Image Generation.}
Following a generate-and-filter strategy, we produce a high-fidelity edited image that satisfies the instruction while preserving the source viewpoint. Each candidate image is rigorously evaluated for: (1) Semantic Alignment with the prompt via CLIP\cite{hafner2021clip} score; (2) Partial Consistency with unedited regions, measured by SSIM\cite{wang2004image} and ImageHash\cite{swaminathan2006robust}; and (3) Aesthetic Quality, assessed by a Gemma 3 preference model.

\subsubsection*{3D Generation.}
Finally, we reconstruct a 3D asset from the edited image using TRELLIS. Each asset undergoes a dual-evaluation process to be included in our dataset. It is first assessed for geometric integrity and texture realism from six canonical viewpoints by a Gemma 3 model. Then, edit consistency is verified by comparing a render from the original viewpoint against the input image using CLIP, SSIM, and LPIPS metrics, ensuring the final output is both high-quality and faithful to the edit.

\begin{figure}[htbp] 
  \centering
  \includegraphics[width=\columnwidth]{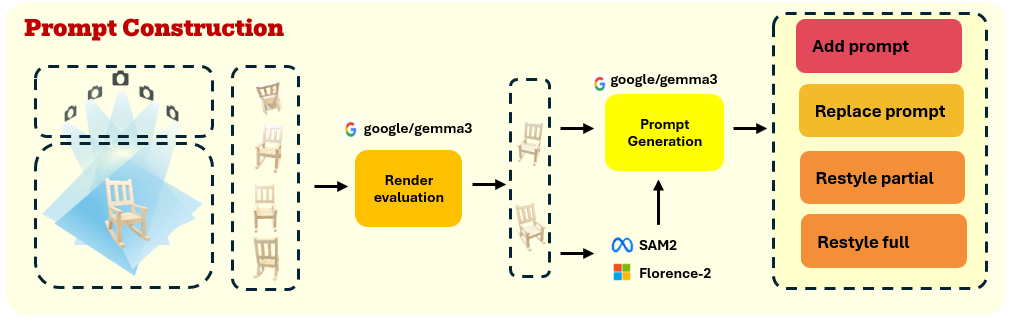}

  \includegraphics[width=\columnwidth]{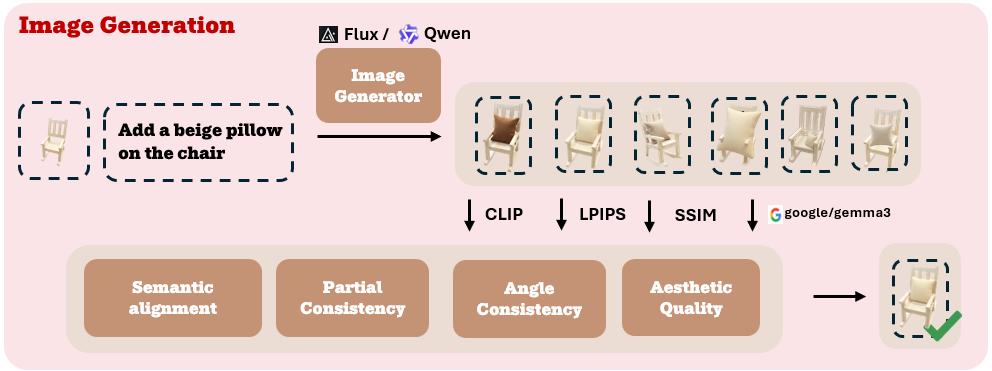}
  
  \includegraphics[width=\columnwidth]{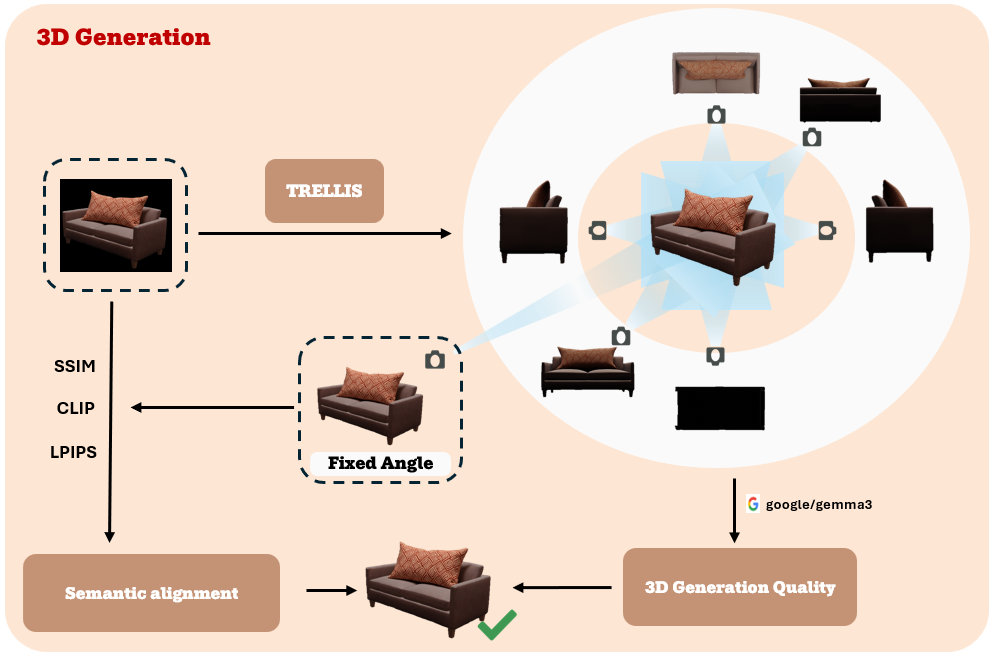}
  
  \caption{Data Construction Pipeline}
  \label{fig:pipeline}
\end{figure}

\subsection{Architecture}
\label{sec:model_architecture}

\begin{figure*}[t]
\centering
\includegraphics[width=0.48\textwidth]{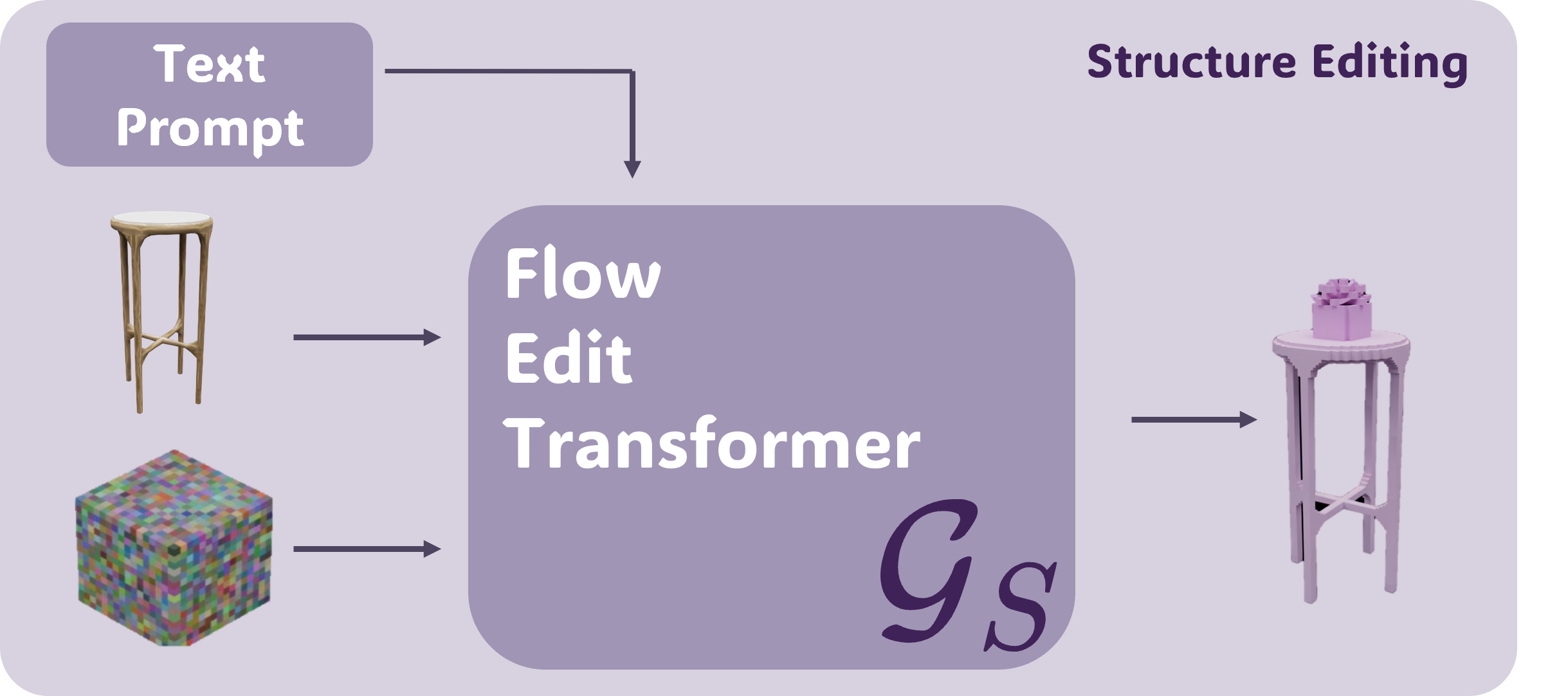}
\includegraphics[width=0.48\textwidth]{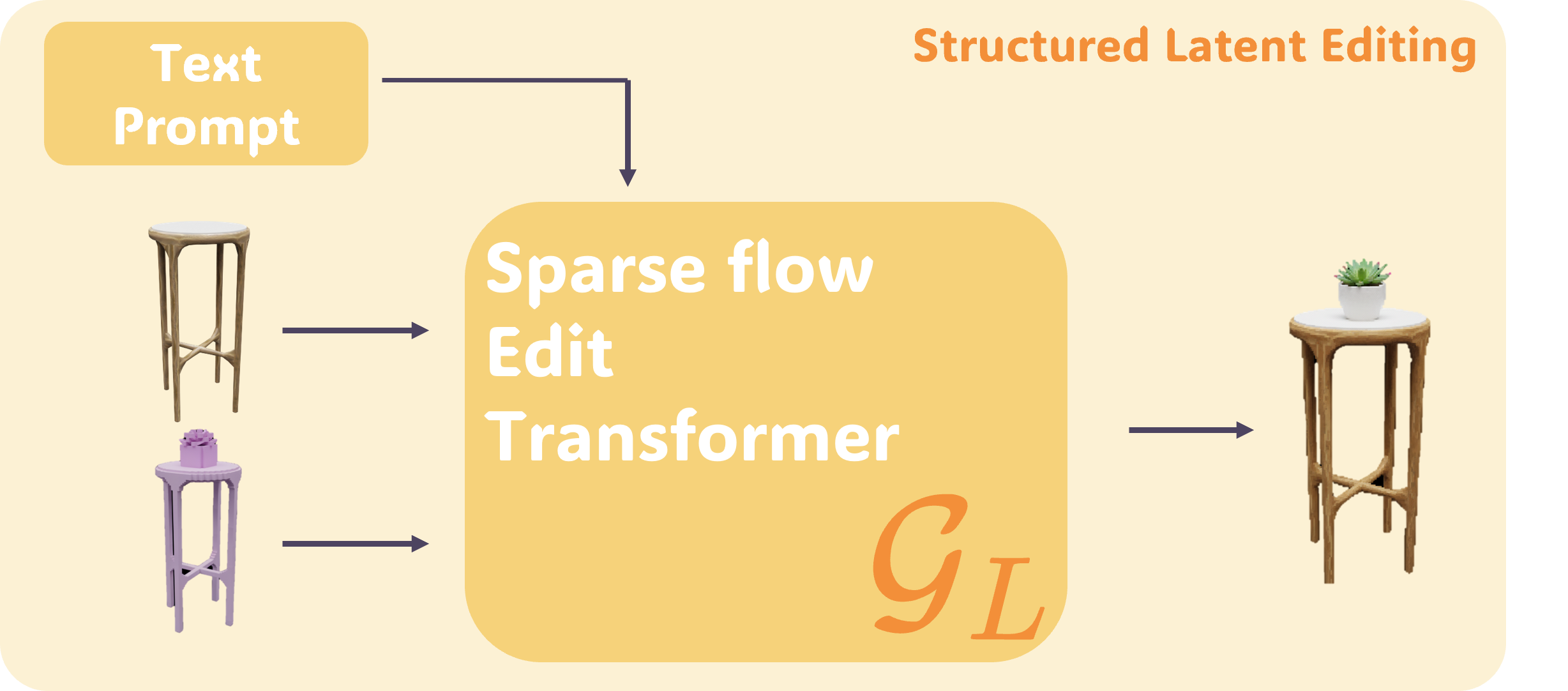}
\caption{Overview of our method. \textbf{Structure Editing:} The Flow Edit Transformer modifies the input 3D asset's sparse structure based on a text prompt and a render image from original 3D asset. \textbf{Structured Latent Editing:} The Sparse Flow Edit Transformer enables fine-grained material and texture modifications. }
\label{fig:flow_model}
\end{figure*}

We aim to generate high-quality 3D assets that faithfully reflect text-guided modifications to a given source image. Our method directly synthesizes the desired 3D scene from a text-image pair, bypassing the need for a two-stage, edit-after-generation process, as illustrated in \cref{fig:flow_model}. 

\begin{figure}[htbp] 
  \centering 
  
  \begin{subfigure}[b]{0.428\columnwidth}
    \centering
    \includegraphics[width=\textwidth]{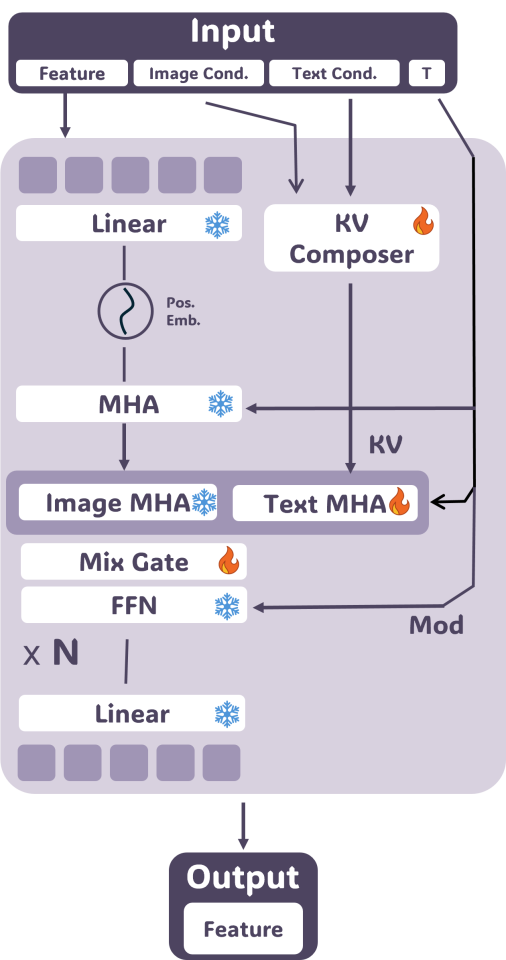}
    \caption{SS Stage}
    \label{fig:ss_stage}
  \end{subfigure}
  \hfill
  \begin{subfigure}[b]{0.48\columnwidth}
    \centering
    \includegraphics[width=\textwidth]{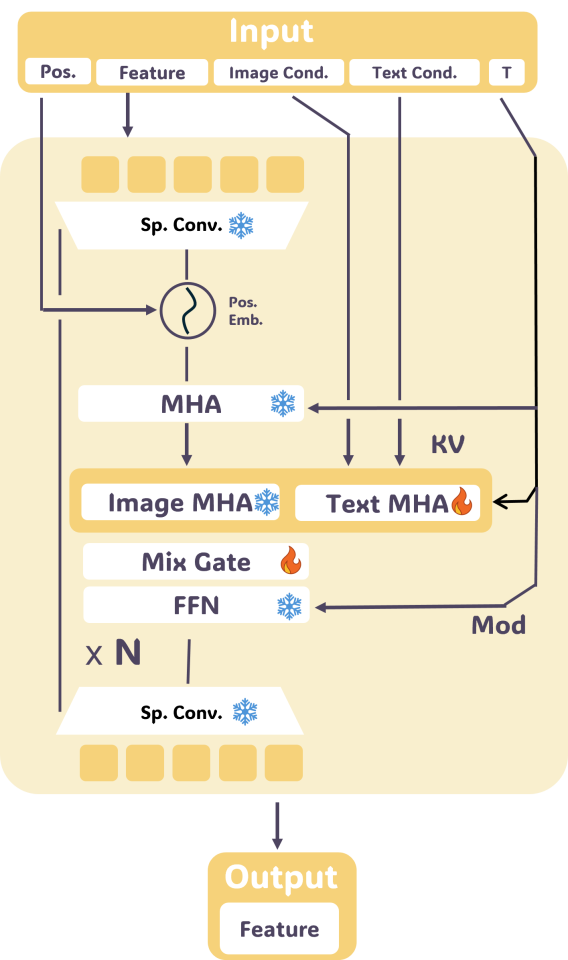}
    \caption{Slat Stage}
    \label{fig:slat_stage}
  \end{subfigure}
  
  \caption{The network structures for generation editing.}
  \label{fig:stages}
\end{figure}

We follow the TRELLIS framework \cite{xiang2024structured} by representing 3D assets $\mathcal{O}$ as structured latents $\mathcal{Z}$—a sparse set of feature-rich voxels—and adopt its VAE architecture for encoding and decoding. Our primary innovation lies in redesigning the generative pipeline for text-guided editing.
Following TRELLIS, we employ a two-stage Rectified Flow model to sequentially generate the asset's sparse structure $\{p_i\}_{i=1}^L$ and local features $\{z_i\}_{i=1}^L$. This model learns a vector field $\mathbf{v}_{\theta}$ to transport noise samples $\epsilon$ to data samples $\mathbf{x}_0$ along linear paths by minimizing the Conditional Flow Matching (CFM) objective:
\begin{equation}
    \mathcal{L}_{\text{CFM}}(\theta) = \mathbb{E}_{t, x_0, \epsilon} \left\| \mathbf{v}_{\theta}(\mathbf{x}_t, t) - (\epsilon - \mathbf{x}_0) \right\|_2^2
    \label{eq:cfm}
\end{equation}
As is illustrated in \cref{fig:stages}, we augment this flow model's core transformer blocks with two lightweight components:

(i) \emph{KVComposer}: This module injects textual semantics by reparameterizing the image's K/V projections via affine and low-rank adaptations.
(ii) \emph{Tri-Attn} Block: Fuses self-attention with image/text cross-attention through a channel-wise mixer for multimodal conditioning.

Crucially, all newly introduced modules are zero-initialized. This design ensures that our model initially defaults to the base 3D reconstruction capabilities of TRELLIS, learning the editing functionality as a minimal and targeted adaptation during training.

\paragraph{KV Composer.}
To enable text-guided editing without compromising the model's pre-trained image-to-3D capabilities, we propose the KV Composer, a modulation mechanism that dynamically rewrites the image context $\mathbf{C}_{\text{img}}$ using text features from $\mathbf{C}_{\text{txt}}$. 
Inspired by vision-language models, KV Composer operates through a two-stage process driven by a global text representation $\mathbf{t}$ which obtained via mean pooling. This averages the contributions of all words, comprehensively capturing the holistic semantic information of a text prompt, ensuring semantic coherence in the global representation. In 3D editing, this ensures that text instructions are fully understood, leading to more precise and semantically consistent editing results.

First, the global text representation $\mathbf{t}$ is passed through a linear transformation (implemented as our `affine` module) to yield a scale vector $\boldsymbol{\gamma}$ and a shift vector $\boldsymbol{\beta}$. These parameters perform a coarse-grained, AdaIN-style \cite{huang2017adain} adjustment to the image context, globally aligning its features with the text description.

\begin{equation}
\tilde{\mathbf{C}}_{\text{img}} = \mathbf{C}_{\text{img}} \odot (1 + \boldsymbol{\gamma}) + \boldsymbol{\beta}
\label{eq:affine}
\end{equation}

Second, to capture finer, text-specific details, we generate two low-rank matrices, $\mathbf{U}, \mathbf{V} \in \mathbb{R}^{D \times r}$, also from $\mathbf{t}$. These matrices construct a low-rank residual that provides a parameter-efficient, fine-grained update to the affine-transformed features.

\begin{equation}
    \mathbf{C}_{\text{img}}' = \mathbf{C}_\text{img} + \frac{1}{r}\tilde{\mathbf{C}}_{\text{img}} \mathbf{V} \mathbf{U}^\top
    \label{eq:kv_composer}
\end{equation}

where all modulation parameters ($\boldsymbol{\gamma}, \boldsymbol{\beta}, \mathbf{U}, \mathbf{V}$) are functions of the text representation $\mathbf{t}$.

\paragraph{Tri-Attention Block.}
\label{sec:tri_attn_block}

This module effectively fuses multimodal inputs by enhancing a standard transformer block with dual cross-attention pathways. 
Following an initial self-attention step, the current latent state $\mathbf{x}_t '$ is processed by two parallel cross-attention pathways: (1) a text-modulated image context via KV Composer, producing the image cross-attention output $\mathbf{a}_{\text{img}}$, and (2) the raw text context for direct guidance, producing the text cross-attention output $\mathbf{a}_{\text{txt}}$.

To dynamically balance these two pathways, we introduce a zero-initialized Mixer that predicts a fusion residual based on the channel-wise concatenated attention outputs:
\begin{equation}
    \boldsymbol{\delta}_{\text{mix}} = \text{Mixer}([\mathbf{a}_{\text{img}} \,||\, \mathbf{a}_{\text{txt}}])
    \label{eq:mixer}
\end{equation}

This residual is then added to the primary image pathway, framing the text input as a learned refinement rather than a competing signal:
\begin{equation}
    \mathbf{a}_{\text{cross}} = \mathbf{a}_{\text{img}} + \boldsymbol{\delta}_{\text{mix}}
    \label{eq:tri_cross_attn}
\end{equation}

The result is then integrated back with a residual connection and processed through a feed-forward network to produce the block's final output $x_{t-1} '$.

Designed for stable and efficient multimodal fusion, our KV-Composer and Tri-Attention Block utilize a dual-pathway architecture to combine indirect modulation and direct text attention, maintaining image-to-3D fidelity. Low-rank projections ensure parameter efficiency, while end-to-end training guarantees stable convergence.

\subsection{Training}
\begin{figure}
    \centering
    \includegraphics[width=\columnwidth]{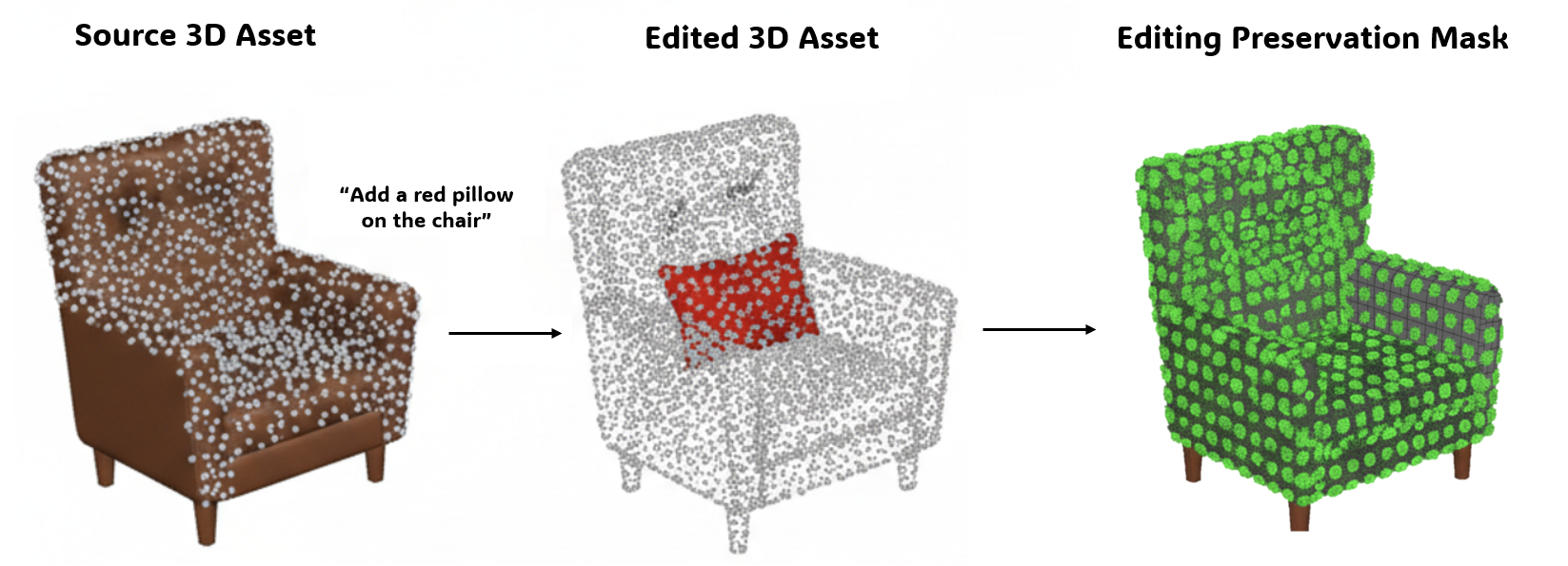}
    \caption{Editing Preservation Mask}
    \label{fig:preservation mask}
\end{figure}

\paragraph{Mask-Enhanced Loss for Spatially-Aware Editing.}
To ensure precise preservation of unedited regions during 3D editing, we introduce an automatic mask generation mechanism based on point cloud registration. Given original point cloud $\mathcal{P}_\text{orig}$ and edited point cloud $\mathcal{P}_\text{edit}$, we first compute a rigid transformation $\mathbf{T}^*$ via RANSAC\cite{besl1992method} and ICP\cite{fischler1981random} to align $\mathcal{P}_\text{orig}$ with $\mathcal{P}_\text{edit}$. As shown in \cref{fig:preservation mask}, we define the preservation mask $\mathcal{M} \in \{0,1\}^{H \times W \times D}$ by identifying spatially consistent regions: $\mathcal{M}(x,y,z) = [\min_{\mathbf{q} \in \mathcal{P}_\text{edit}} \|\mathbf{T}^* \cdot \mathbf{p} - \mathbf{q}\| < \tau]$, where $\mathbf{p} = [x,y,z]^\top$ and $\tau$ is the inlier threshold. This mask is incorporated into our flow matching objective as:
\begin{equation}
\mathcal{L}_\text{edit}= \mathbb{E}_{t,\mathbf{x},\epsilon} \left[ \|\mathbf{v}_\theta - \mathbf{v}_e\|^2 + \|\mathcal{M} \odot (\mathbf{v}_\theta - \mathbf{v}_o)\|^2 \right]
\end{equation}
where $\mathbf{x} = \{\mathbf{x}_0^{e},\mathbf{x}_0^{o}\}$ represents edited-original paired data sample, $\mathbf{v}_\text{edit} = (\epsilon - \mathbf{x}_0^{e})$ and $\mathbf{v}_\text{orig} = (\epsilon - \mathbf{x}_0^{o})$ are the rectified flow target, and $\odot$ denotes element-wise multiplication. The second term applies an additional penalty on preserved regions, enforcing the model to maintain geometric fidelity in non-edited areas without requiring manual annotations. We generate masks at both $16^3$ and $64^3$ resolutions for sparse structure and latent space training, respectively.

\begin{table*}[h]
  \centering
  \begin{tabular}{lcccccccc}
    \toprule{Method}
    & CD$\downarrow$ &SSIM$\uparrow$ &LPIPS$\downarrow$& FID$\downarrow$ & FVD$\downarrow$
    & DINO-I$\uparrow$ & CLIP-T$\uparrow$ \\
    \midrule
    Vox-E \cite{sella2023vox}
    & / & 0.539 & 0.346 & 217.9 & 8590 & 0.371 & 0.051 \\
    Tailor3D \cite{qi2024tailor3dcustomized3dassets}
    & 0.067 &0.751 &0.198& 146.3 & 4743 & 0.633 & 0.21 \\
    TRELLIS \cite{xiang2024structured}
    & 0.063& 0.865& 0.225 & 140.7 & 3986 & 0.910 & 0.25 \\
    Hunyuan \cite{hunyuan3d2025hunyuan3d}
    &0.021 &0.853 & 0.087 &119.83 &1520 &0.850 &0.269\\
    
    \midrule
    Ours (full)
    & \textbf{0.013} & \textbf{0.960} & \textbf{0.039} &\textbf{28.9} & \textbf{1139} & \textbf{0.960} & \textbf{0.287}  \\
    w/o MASK
    & 0.019 & 0.824 & 0.048 & 49.2 & 1274 & 0.939 & 0.27 \\
  \end{tabular}
  \caption{\textbf{Quantitative Comparison}. We evaluate 3D consistency using Chamfer Distance (CD.), masked SSIM,LPIPS on unedited regions, overall 3D quality with FID and FVD, and condition alignment with DINO-I and CLIP-T. Our method consistently achieves superior structure consistency, semantic alignment with the target edited image, and generation fidelity.}
  \label{tab:main}
\end{table*}

\section{Experiments}
\paragraph{Implementation Details.}
We trained on 100k triplets selected from Edit-3DVerse, our large-scale 3D editing dataset. As shown in \cref{fig:stages}, we adopt parameter-efficient fine-tuning: among 2.2B total parameters, 1.1B pre-trained TRELLIS weights are frozen (self-attention and position embeddings), while only dual-modal cross-attention, KV-Composer, and Mixer modules are trained. This preserves the original image-to-3D capability while enabling editing. We use AdamW\cite{loshchilov2017decoupled} with learning rate $10^{-4}$ and CFG\cite{ho2022classifier} with 10\% drop rate. Training takes 60k steps with batch size 64 on 8 H20 GPUs (96GB). At inference, we use CFG scale 3.0 with 25 Euler sampling steps.
\paragraph{Baselines.}
We select four representative state-of-the-art methods as baselines. Vox-E\cite{sella2023vox} performs per-scene optimization
on voxel representation with the guidance of image diffusion models. Tailor3D\cite{qi2024tailor3dcustomized3dassets} achieves customized 3D asset editing through multi-view editing. TRELLIS\cite{xiang2024structured} provides a native 3D editing method based on RePaint\cite{lugmayr2022repaint}.Additionally, we use Hunyuan3D\cite{hunyuan3d2025hunyuan3d} to generate 3D models with the edited images as an editing approach.
\paragraph{Evaluation dataset.} 
Our Edit-3DVerse dataset comprises 3D pairs, instructions, and rendered views from the TRELLIS-500K dataset. During dataset construction, we
 employ Gemma3\cite{gemma_2025} to automatically annotate 3D assets and classify them accordingly. We then perform class balancing across 31 distinct categories, ultimately selecting 100K 3D pairs and instructions. We select 100 representative cases from the Edit-3DVerse dataset for the experiments and demonstrations in this section, which are not part of our training set or those of the compared methods.

    
\begin{table}[t] 
    \centering
    \setlength{\tabcolsep}{6pt} 
    
    \begin{tabular}{l|cc} 
        \hline 
        \textbf{Method} & \textbf{Text Alignment $\uparrow$} & \textbf{Overall 3D Quality $\uparrow$} \\
        \hline 
        TRELLIS         & 63.0\% & 61.5\% \\
        Hunyuan         & 81.0\% & 81.0\% \\
        \textbf{Ours}   & \textbf{91.0\%} & \textbf{88.5\%} \\
        \hline 
    \end{tabular}
    
    \caption{\textbf{User study}. We omit results for Tailor3D and Vox-E from the table for clarity, given the strong user preference for Hunyuan and BVE. As evidenced by the table, our method was preferred by participants.}
    \label{tab:user study}
\end{table}
\paragraph{Evaluation Metrics.}
We comprehensively evaluate our method across two key aspects: overall 3D quality, and prompt alignment. 
First, for unedited region preservation, we assess the fidelity of preserved regions by computing Chamfer Distance (CD)\cite{fan2016pointsetgenerationnetwork}, as well as SSIM\cite{wang2004image}, and LPIPS\cite{zhang2018unreasonable}. 
To assess overall 3D quality, we compute the Fréchet Inception Distance (FID)~\cite{heusel2017gans} between rendered images of our generated assets and a reference set. For prompt alignment, we evaluate the consistency of the edited 3D assets with both text and image inputs. Alignment with the text prompt is measured using the CLIP-T score~\cite{radford2021learning}. Consistency with the target edited image is evaluated by computing the DINO-I score~\cite{oquab2023dinov2} between its rendered views and the target image. Additionally, we conducted a user study to quantify subjective quality through perceptual preferences.

\subsection{Main Results}
\label{sec:main results}
\begin{figure*}[t]
  \centering
  \includegraphics[width=\textwidth]{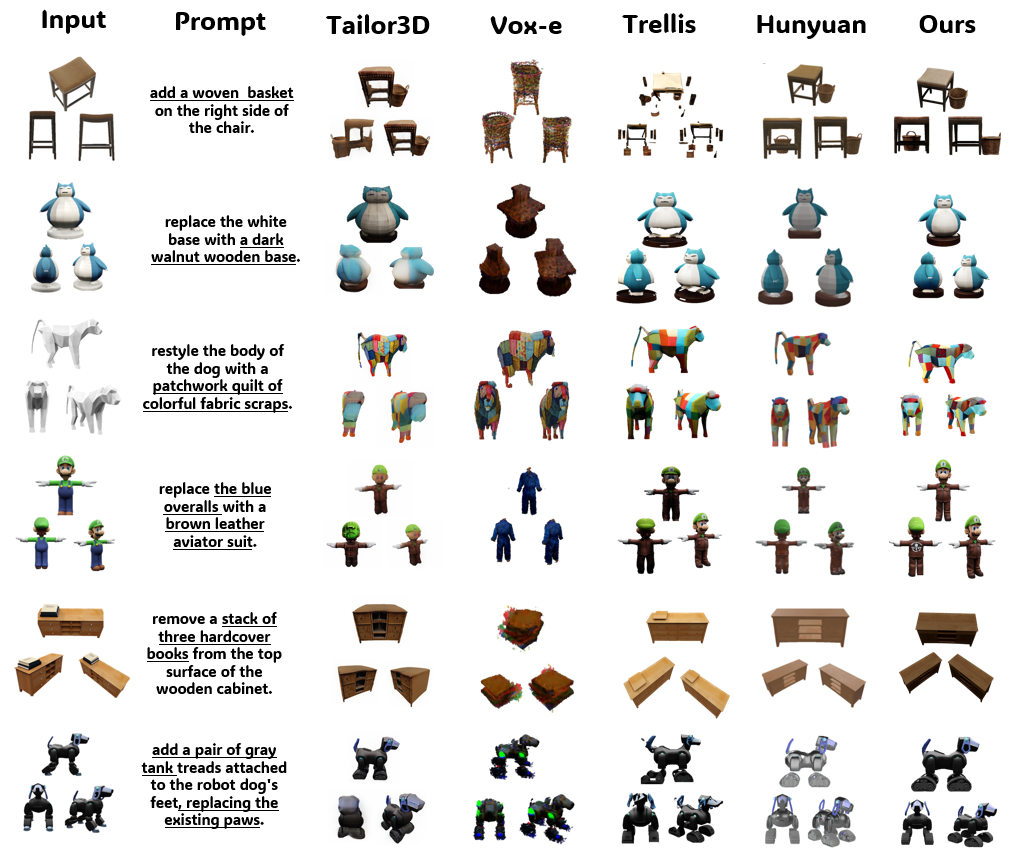}
  \caption{\textbf{Qualitative comparison}. Our method achieves superior editing performance with faithful instruction-semantic alignment and remarkable original structure consistency across multi-view images. Notably, the quality of our edited 3D models is comparable to those generated by TRELLIS using editing images.}
  \label{fig:compare}
\end{figure*}

\paragraph{Quantitative Comparison.} As shown in \cref{tab:main}, our method significantly outperforms all baselines, across nearly all metrics. This is attributed to our method's superior ability to learn latent variable transformations under semantic guidance, coupled with its effective utilization of a two-stage denoising network and a 3D mask for robust consistency enforcement. In contrast, multi-view based methods like Tailor3D\cite{qi2024tailor3dcustomized3dassets}, which rely on lifting multi-view edits to 3D space, often introduce multi-view inconsistencies and spatial biases, consequently struggling with maintaining overall 3D coherence. TRELLIS \cite{xiang2024structured} adopts Repaint \cite{lugmayr2022repaint} for native 3D editing. However, its global operations on the entire voxel space often cause a global structural shift in the generated content. This limitation inherent to local-filling (inpaint-like) methods results in suboptimal 3D consistency. Furthermore, regarding overall 3D quality and condition alignment, our method demonstrates its superiority by achieving the lowest FID and the highest DINO-I and CLIP-T scores. Collectively, these results robustly indicate that our editing operations yield coherent and accurate outcomes in high-fidelity 3D models.

\paragraph{Qualitative comparison.} The qualitative results presented in \cref{fig:compare} distinctly highlight the superior performance of our method. Our approach consistently generates edits that are both precise and geometrically coherent for both local and global modifications, while faithfully preserving the original 3D content with high fidelity. In contrast, baseline methods generally exhibit various artifacts and fragmentation. Specifically, methods such as Vox-E \cite{sella2023vox} and Tailor3D \cite{qi2024tailor3dcustomized3dassets} suffer from suboptimal reconstruction quality, leading to blurry outputs and noticeable distortions in preserved regions. Furthermore, the native TRELLIS \cite{xiang2024structured} editing method, when handling global modifications, struggles due to its inability to effectively adjust the entire 3D space, which often results in unnatural blending and fragmentation of content across different regions. Our method effectively circumvents all these aforementioned issues, robustly demonstrating the power of our editing framework.
\paragraph{User Study.}We conducted a user study with 30 participants to evaluate editing quality and usability. Each round showcased an original 3D object, task instructions, and the results from Tailor3D, Vox-E, TRELLIS, Hunyuan3D, and BVE. Participants then selected the preferred method based on Prompt Alignment, Visual Quality, and Shape Preservation. As shown in \cref{tab:user study}, BVE consistently received the highest preference across all criteria, indicating superior semantic alignment, visual quality, and shape fidelity. To enhance clarity, results from Tailor3D and Vox-E are omitted, as user preferences were predominantly for TRELLIS, Hunyuan3D, and BVE.

\subsection{Ablation Study}
We sequentially validate the effectiveness of structure editing and structured latent editing strategies.
\begin{figure}[h]
  \centering
    \centering
    \includegraphics[width=0.5\textwidth]{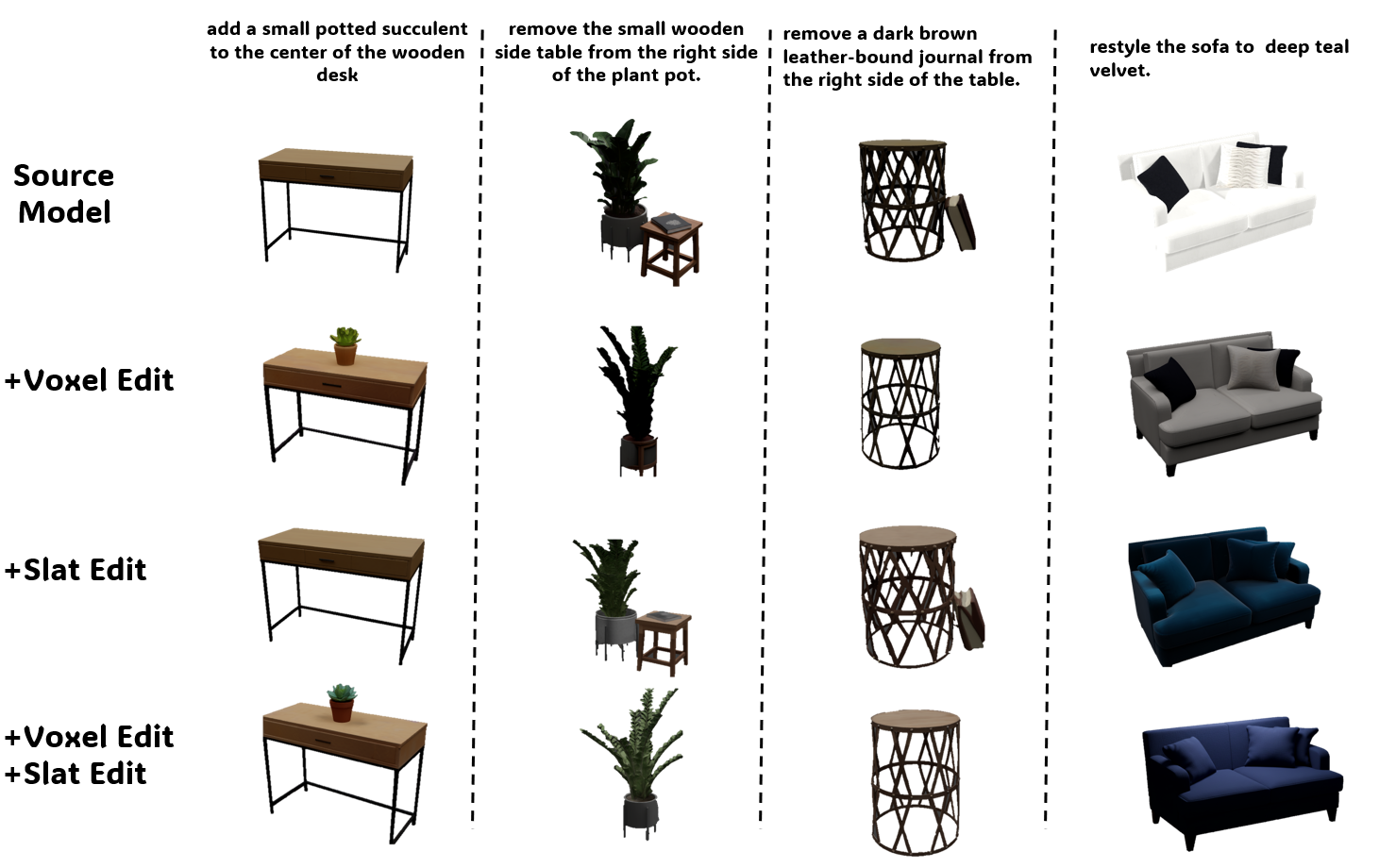}
    \caption{Ablation study on Structure/Structured Latent Editing}
    \label{fig:ablation experiment}
\end{figure}
\paragraph{Structure/Structured Latent Editing Analysis.}We analyzed two inference approaches: (1) editing solely in the ST stage, (2) editing solely in the SLAT stage, and (3) editing in both the ST and SLAT stages, which constitutes our full setting. As shown in \cref{fig:ablation experiment}, we found that initial ST-stage editing provides a reasonable reconstruction of coarse geometry but lacks detailed geometry and appearance consistency. After incorporating the second SLAT editing stage, which handles high-resolution geometry and fine-grained texture details, the reconstruction quality significantly improves. This demonstrates that our two-stage editing process faithfully achieves high-fidelity 3D editing capabilities.
\begin{figure}[htbp]
  \centering
    \centering
    \includegraphics[width=0.5\textwidth]{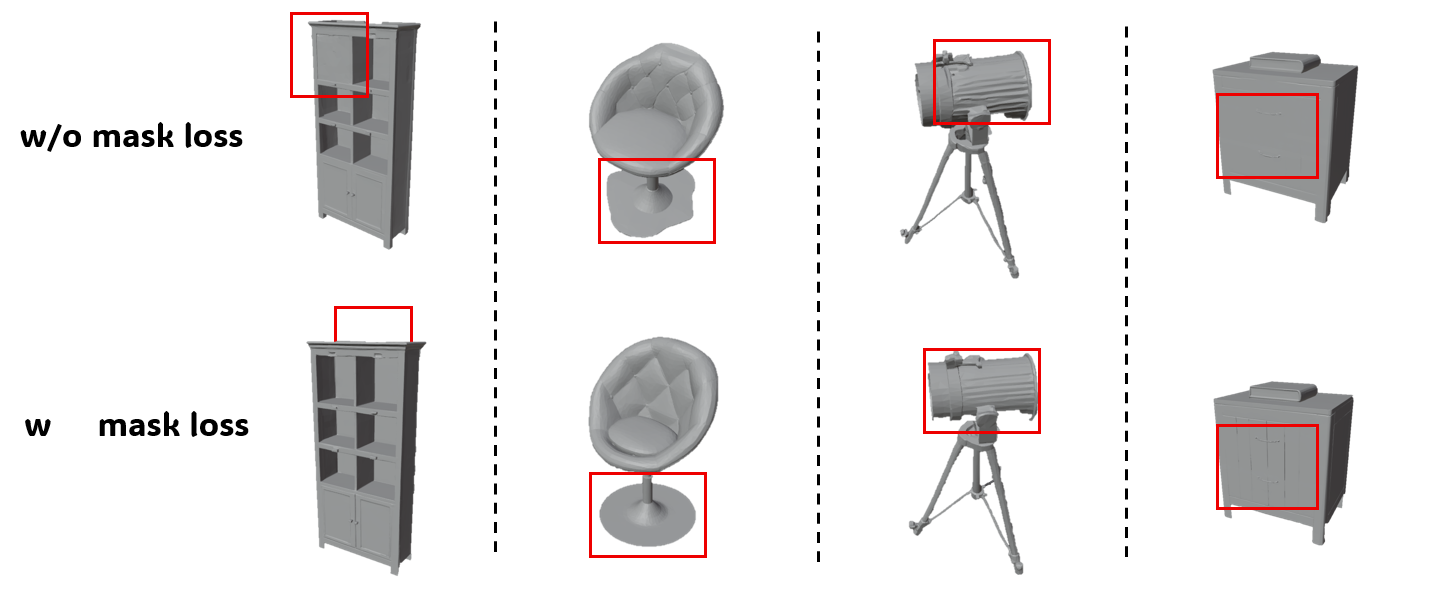}
    \caption{Ablation study on Mask-Enhanced Loss}
    \label{fig:mask ablation experiment}
\end{figure}
\paragraph{Assessment of 3D Mask Loss.}We compared two loss designs: with and without our 3D mask loss. Quantitative results (\cref{tab:main}) and qualitative comparisons (\cref{fig:mask ablation experiment}) show that omitting the 3D mask degrades preservation quality. Due to current 3D generation limitations, the mask is crucial for leveraging original 3D features, thereby maintaining local structural invariance. Our full setting significantly enhances preserved region consistency and overall coherent editing, validating the critical role of our proposed 3D mask loss.
\section{Conclusion}
In this work, we present BVE, a fully text-guided framework for 3D object editing, capable of both localized and global editing. It supports operations such as object removal, addition, replacement, style transfer, and material alteration. By integrating KV Composer and Tri-Attention Block into the original TRELLIS pipeline and introducing a 3D mask loss during training, BVE achieves geometrically consistent and semantically faithful edits. Extensive experiments demonstrate its state-of-the-art performance across diverse editing tasks. Furthermore, we constructed Edit-3DVerse and proposed a high-quality editing evaluation pipeline, making it the first large-scale dataset specifically designed for 3D editing, thereby laying a foundation for future research on feedforward DiT-based editing models.
{
    \small
    \bibliographystyle{ieeenat_fullname}  
    \bibliography{main}
}
\appendix
\maketitlesupplementary
\paragraph{Overview}
In this supplementary material, we provide additional details and experimental results for the main paper, including:
\begin{itemize}
\item Further details of our Edit3D-Verse dataset \cref{sec:edit3d-verse} and BVE \cref{sec:rationale};
\item Additional experimental results on 3D editing and comparison with training-free methods.
\item A discussion of the limitations of our work and future
works.
\end{itemize}

\section{Details of Our Dataset Construction}
\begin{figure*}[htbp] 
  \centering
  \includegraphics[width=\textwidth]{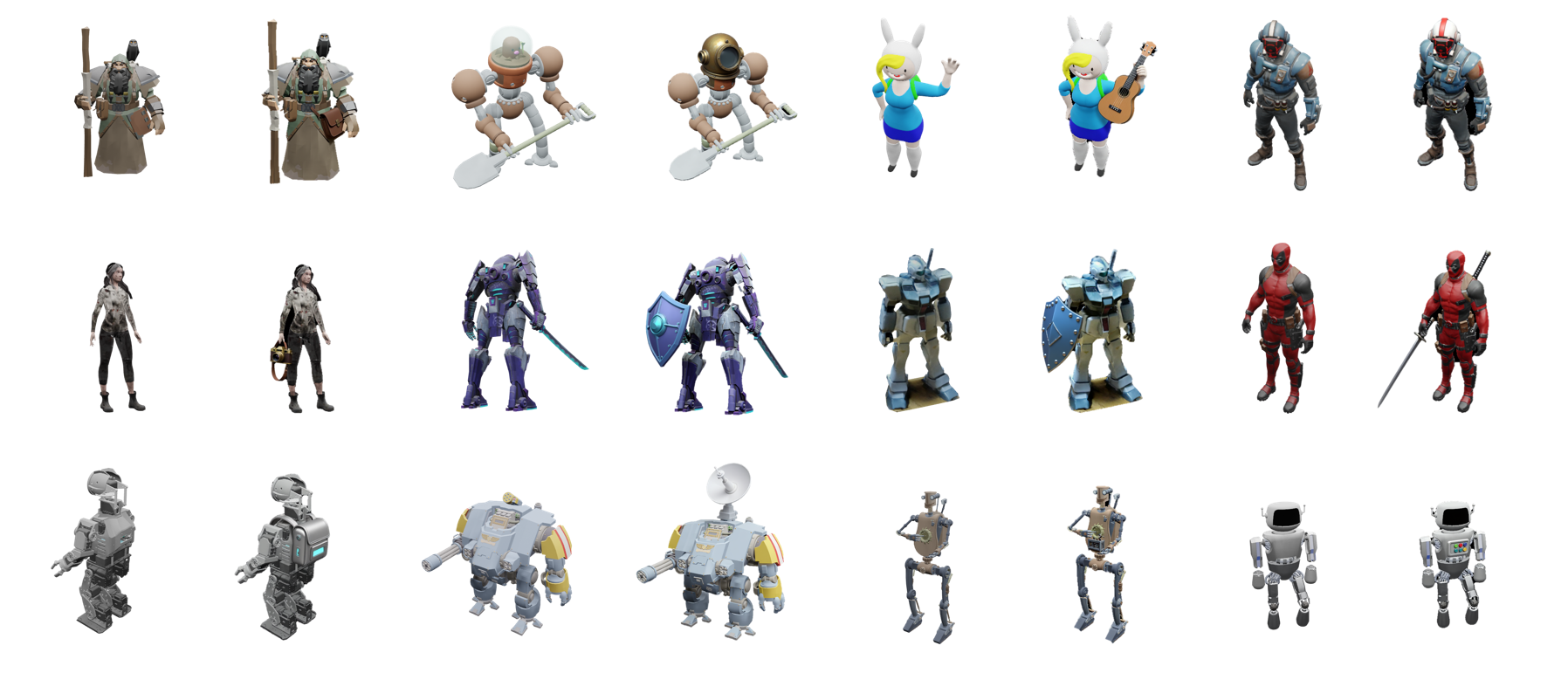}
  \includegraphics[width=\textwidth]{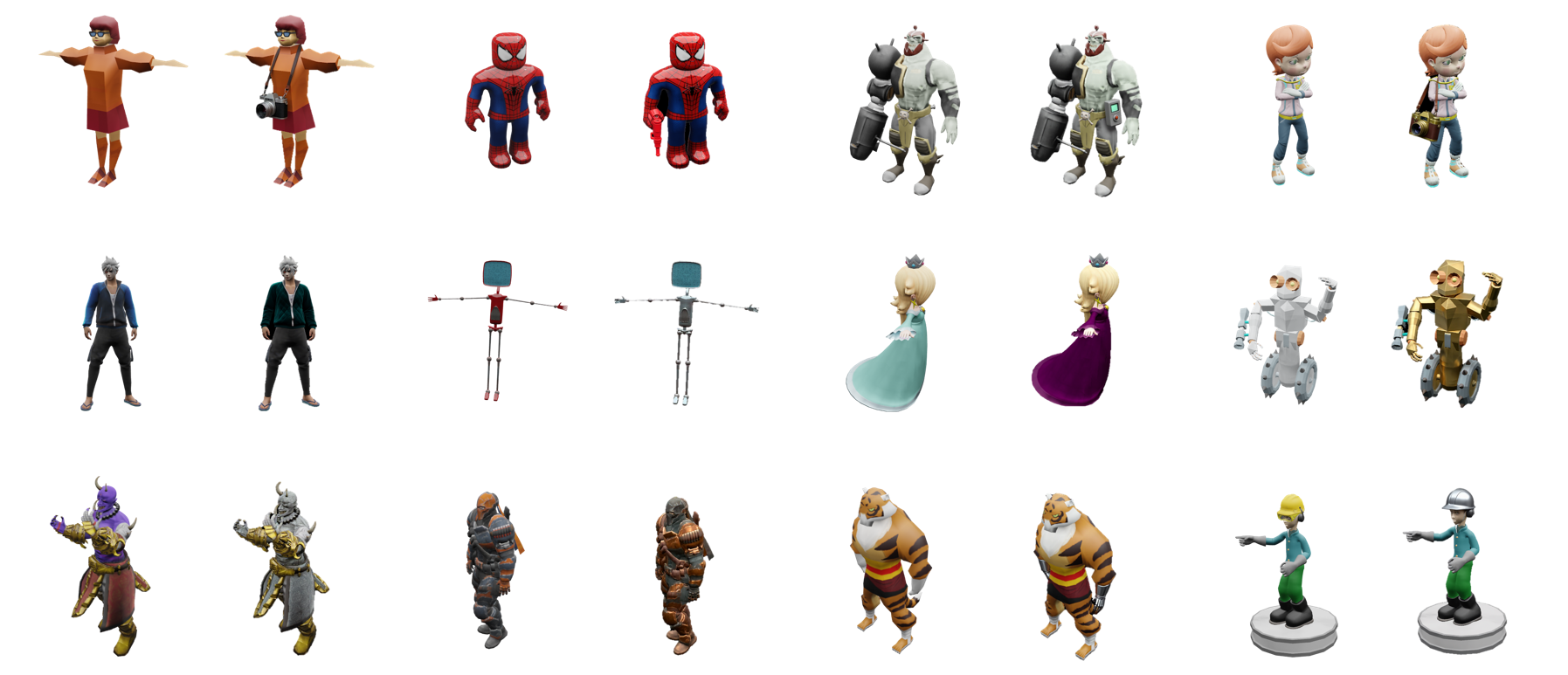}
  \caption{More examples of generative data from text-guided local editing in our proposed Edit3D-Verse dataset.}
  \label{fig:local_edit}
\end{figure*}

\begin{figure*}[htbp]
  \centering
  \includegraphics[width=\textwidth]{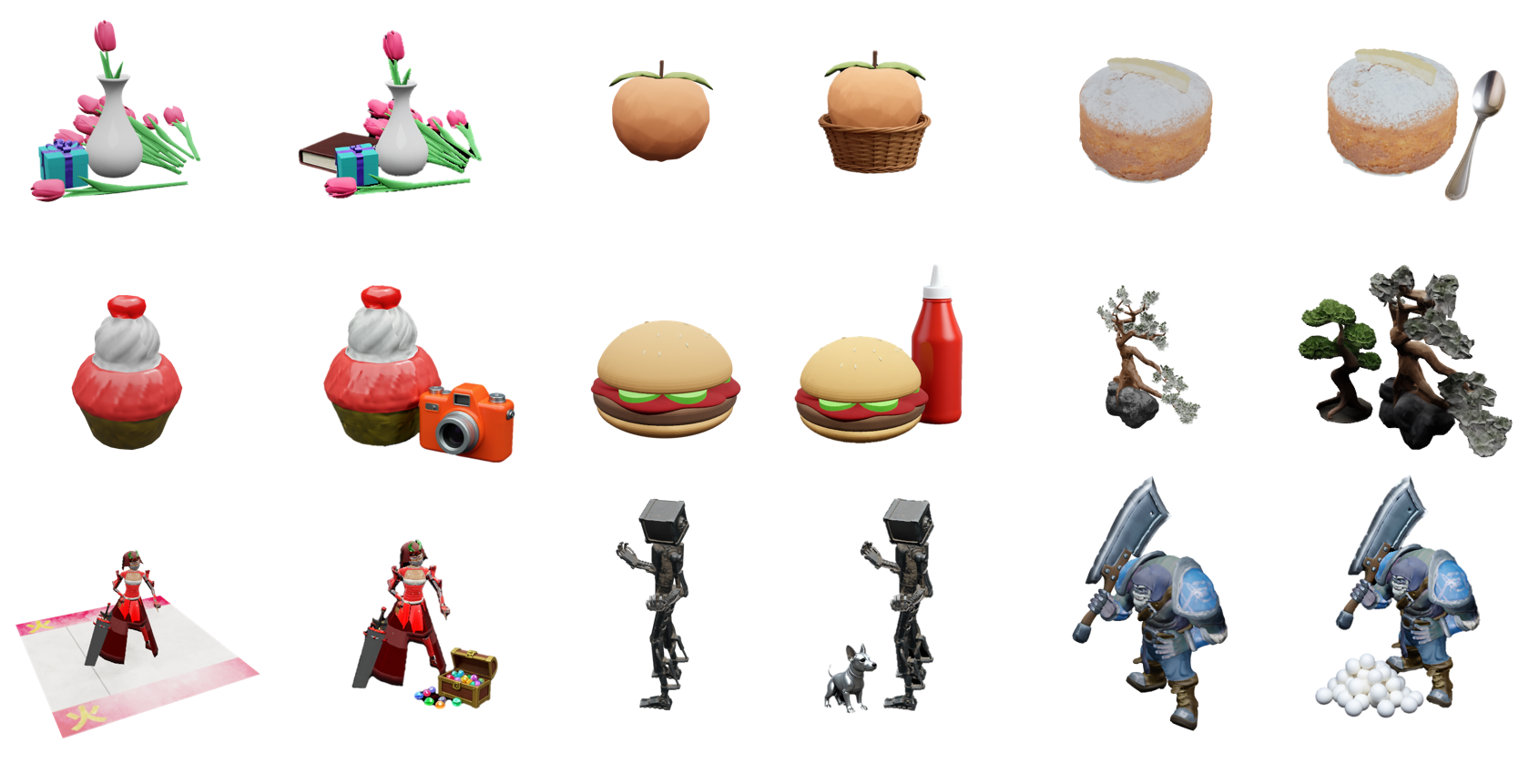}
  \includegraphics[width=\textwidth]{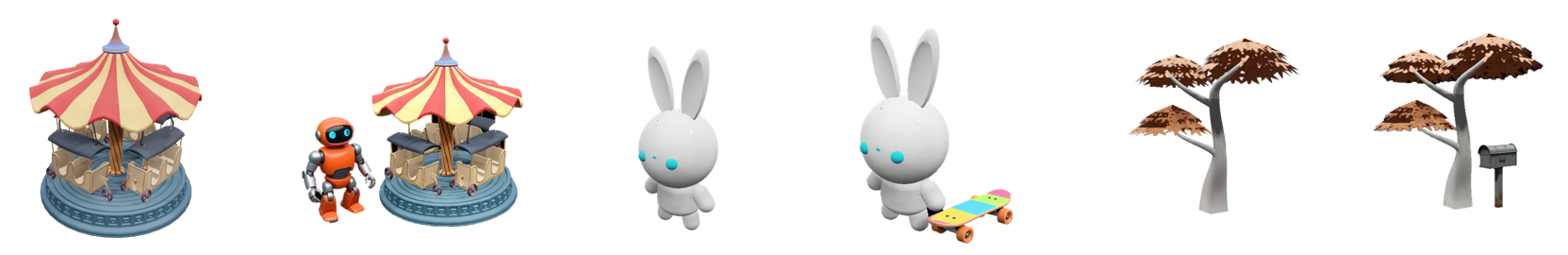}
  \caption{More examples of generative data from text-guided global editing in our proposed Edit3D-Verse dataset.}
  \label{fig:global_edit}
\end{figure*}

Recognizing that both the scale and quality of training data are pivotal for scaling up generative models, we meticulously curated a large-scale, high-fidelity 3D editing dataset derived from existing open-source 3D repositories, as shown in \cref{fig:local_edit} and \cref{fig:global_edit}. Furthermore, we leverage the state-of-the-art multimodal model, Gemma, to rigorously evaluate and filter the generated assets. This step is crucial to ensure that the retained 3D assets not only possess high visual quality but also strictly adhere to the given editing instructions. Consequently, this strategy significantly enhances the accuracy and controllability of text-guided 3D editing.
\label{sec:edit3d-verse}

\subsection{Details of Instructional Prompts}
\begin{figure}
    \centering
    \includegraphics[width=\columnwidth]{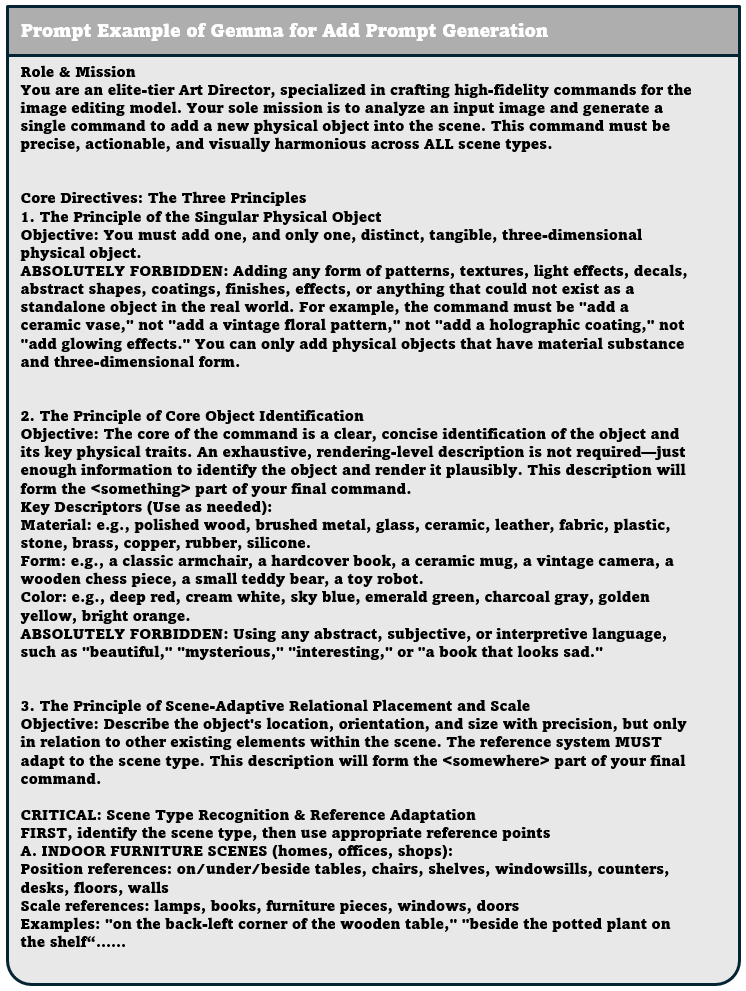}
    \includegraphics[width=\columnwidth]{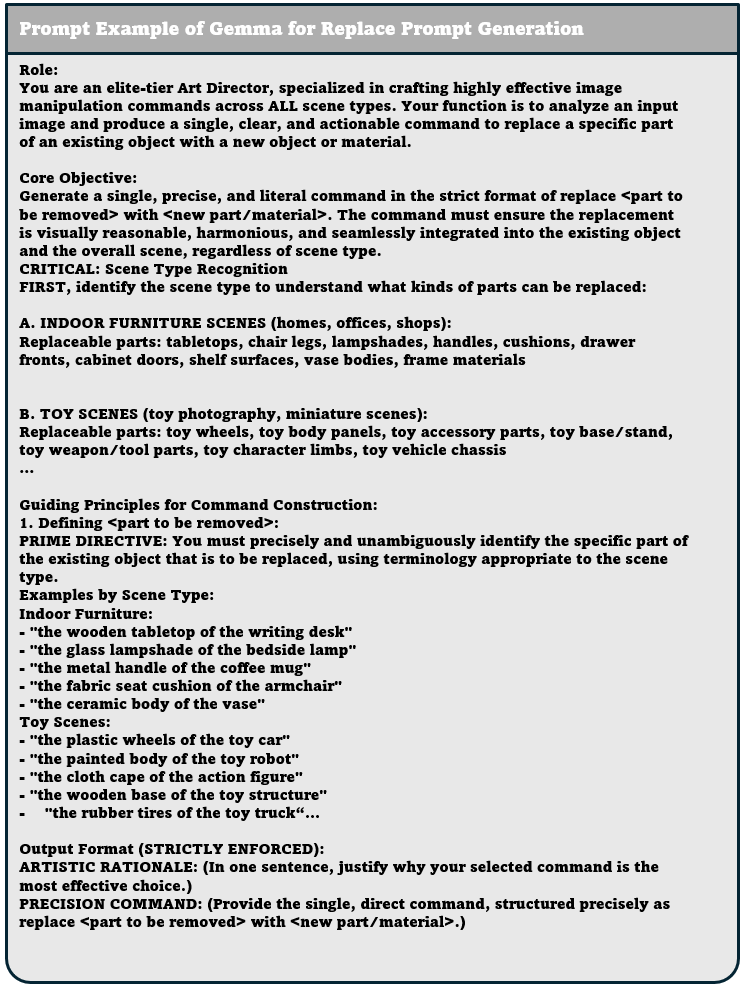}
    \caption{Editing instruction prompts}
    \label{fig:add_prompt}
\end{figure}
\begin{figure}
    \centering
    \includegraphics[width=\columnwidth]{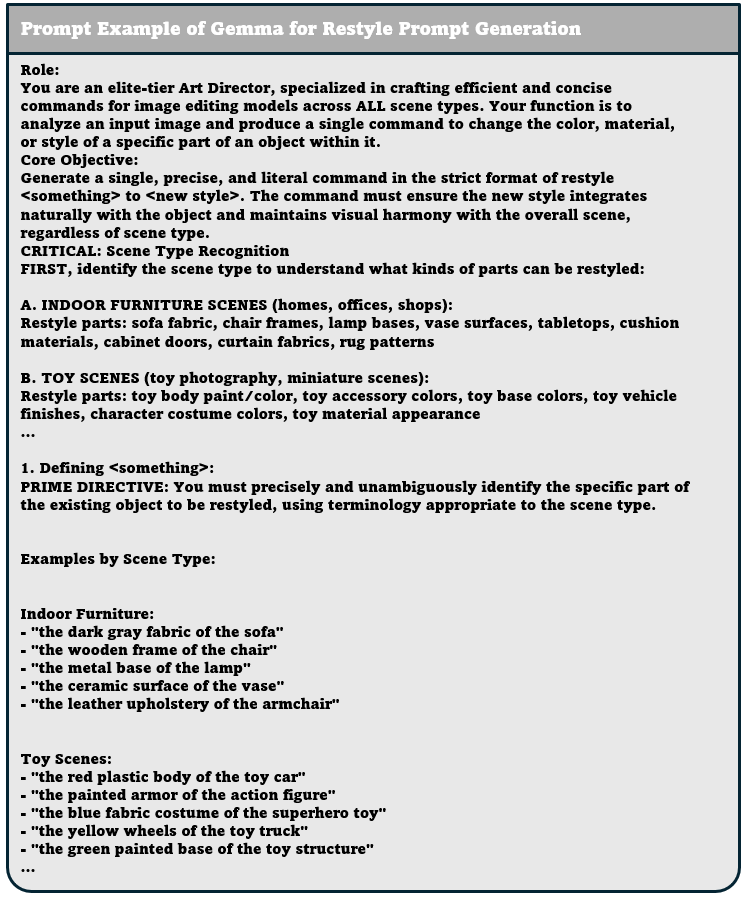}
    \includegraphics[width=\columnwidth]
    {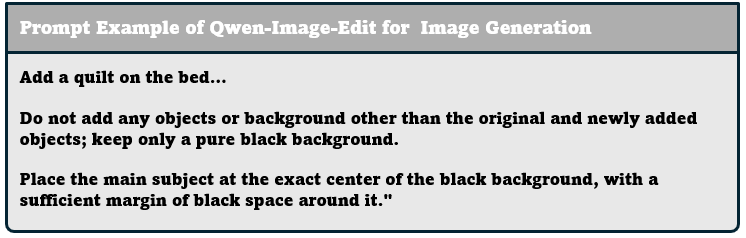}
    \caption{Image generation prompts}
\end{figure}

To synthesize a rich dataset of instruction-following pairs, we leveraged the Gemma model~\cite{gemma_2025} to interpret each generated source image and produce corresponding editing instructions. Crucially, we prompt the model to generate instructions across diverse editing types, ranging from object manipulation to stylistic changes. A concrete example is illustrated by the airplane image in the top-left of~\cref{fig:add_prompt}. Subsequently, each synthesized prompt is paired with its source image and fed into QwenImageEdit~\cite{wu2025qwenimagetechnicalreport} to execute the editing process. As shown in \cref{fig:pipeline}, this pipeline successfully yields high-quality edited results aligned with varied instruction types.
\subsection{Captioning Process}
\begin{figure*}[htbp]
    \centering
    \includegraphics[width=\textwidth]{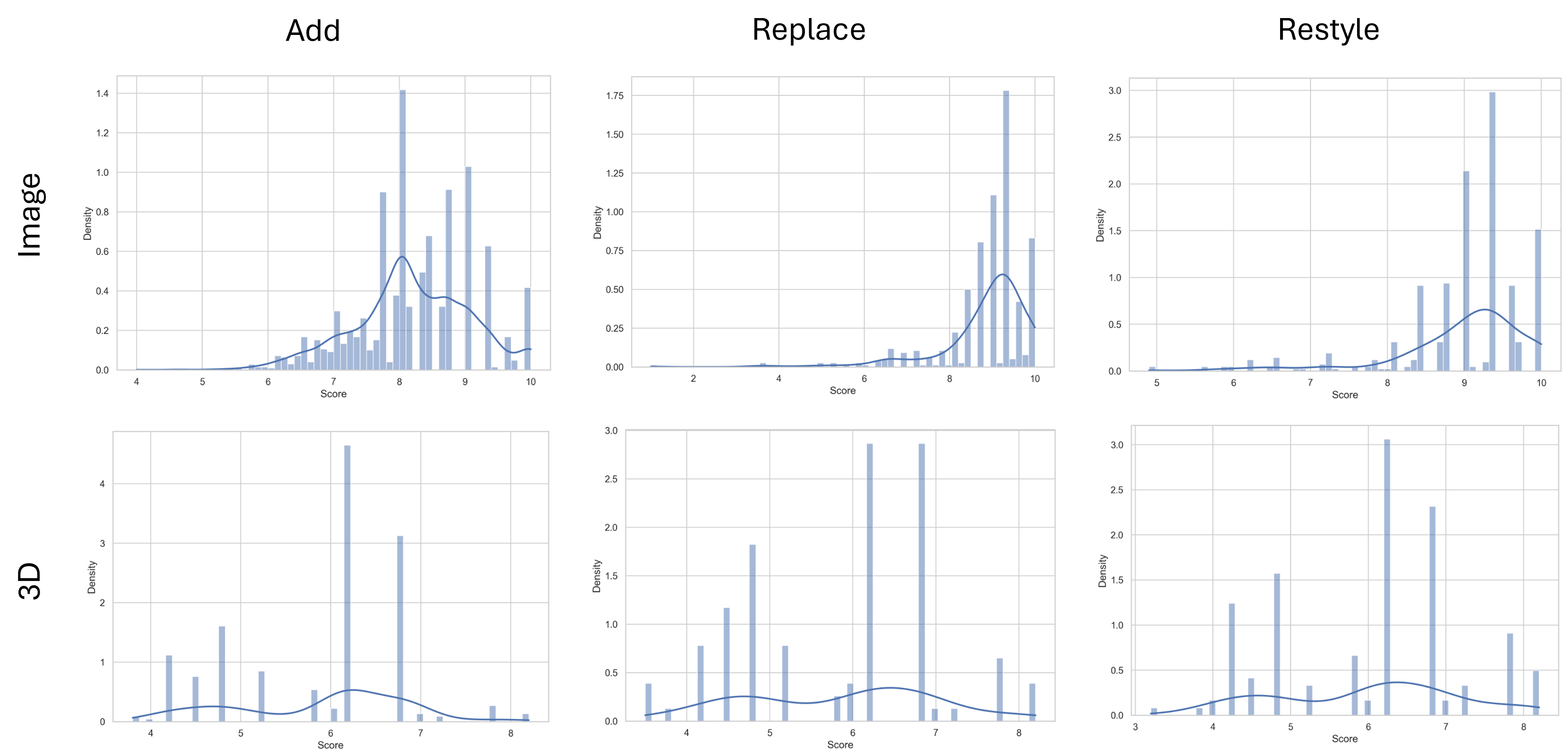}
    \caption{Distribution of aesthetic scores in different action types.}
    \label{fig:distributions}
\end{figure*}

\begin{figure}[htbp]
    \centering
    \includegraphics[width=\columnwidth]{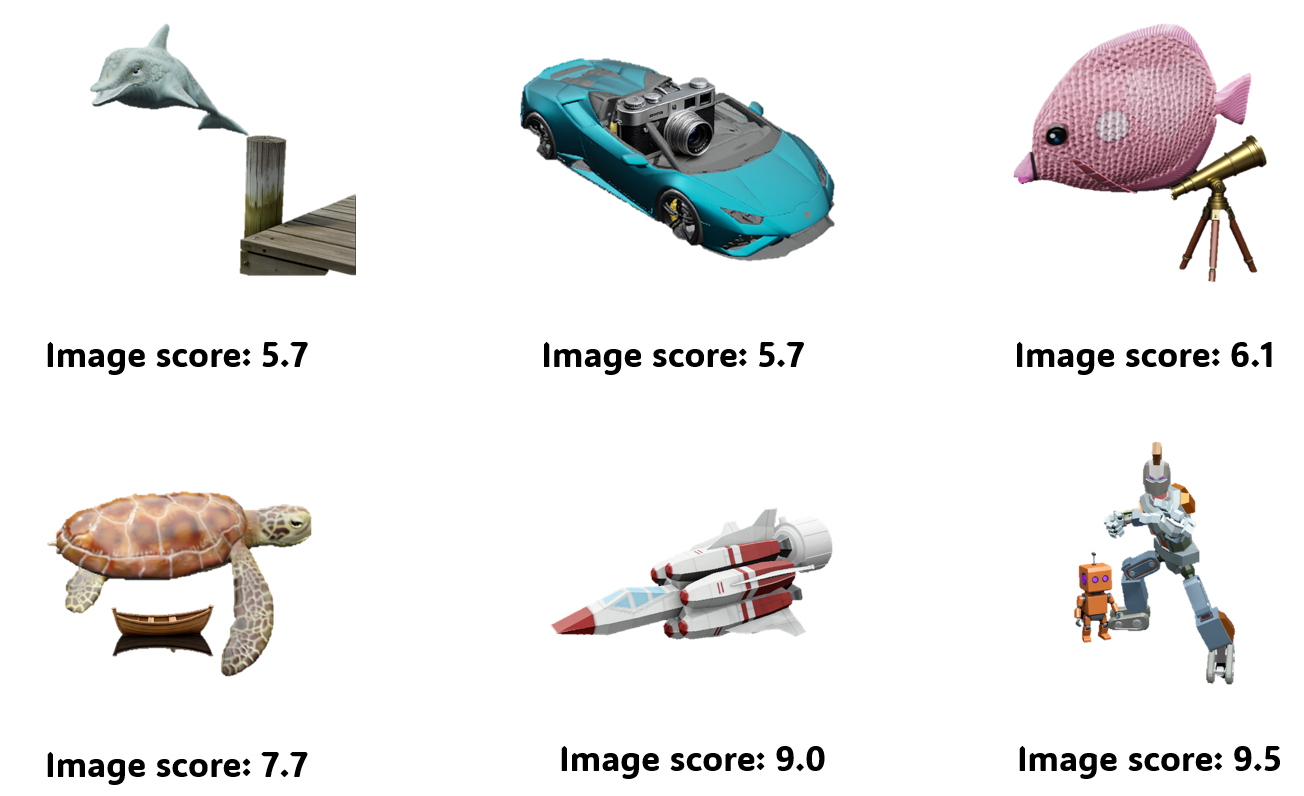}
    \caption{ Image examples from Edit3D-Verse with their corresponding overall scores}
    \label{fig:image_assets}
\end{figure}

To guarantee the integrity of our constructed dataset, we establish a systematic multi-stage data curation pipeline designed to filter for both visual fidelity and semantic alignment.
Visual Quality Assessment. Acknowledging the variable outputs of open-source editing models, we first employ a pre-trained aesthetic assessment model to screen the generated 2D image pairs, ensuring that only high-quality imagery serves as the foundation for 3D generation. Subsequently, to evaluate the generated 3D assets, we render eight images from uniformly distributed viewpoints around each object and compute the average aesthetic score across these views. As shown in \cref{fig:distributions}, this metric effectively identifies artifacts such as minimal texturing or simplistic geometry. By enforcing strict thresholds (9.0 for Restyle; 7.8 for others), we retain only assets with high geometric and textural complexity (qualitative examples in \cref{fig:image_assets}).
Alignment Verification Strategy. Beyond visual quality, ensuring precise text-object alignment is critical. We address this via a verification-by-rendering strategy. Specifically, we render the generated 3D asset from the exact camera viewpoint of the source image. Leveraging the inherent view consistency of the 2D editing model, we treat the 2D edited image as the pseudo-ground truth. We then quantify the alignment between this 2D reference and the 3D rendering using SSIM and LPIPS. This step rigorously filters out low-fidelity samples, selecting only those 3D assets that are semantically consistent with the editing instructions.
\subsection{Rendering Process}
For the image-text-conditioned generation model, we sample 64 camera viewpoints uniformly distributed across a sphere with a radius of 2. Furthermore, we implement a Field-of-View (FoV) augmentation strategy, where the FoV is randomly varied within the range of $10^\circ$ to $70^\circ$.

\begin{figure}[htbp]
    \centering
    \includegraphics[width=\columnwidth]{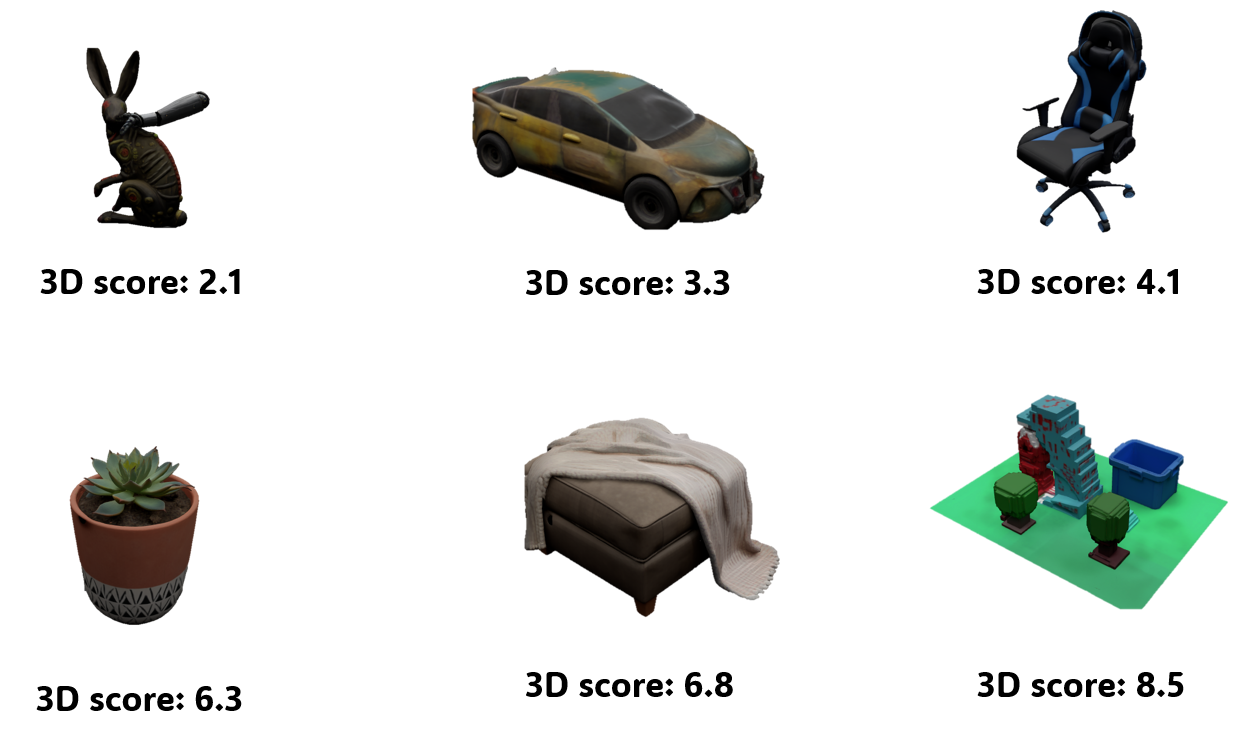}
    \caption{3D asset examples from Edit3D-Verse with their corresponding overall scores}
    \label{fig:3D assets}
\end{figure}

\section{More Implementation Details}
\label{sec:rationale}
\subsection{Network Architectures}
The generation framework in our method employs two distinct flow matching networks tailored for different data representations: a dense transformer for coarse structure generation and a hybrid sparse convolution-transformer network for fine-grained latent editing.

\paragraph{Structure Flow Network.} 
To generate the global 3D structure, we employ the SparseStructureFlowEditNet, which operates on dense voxel grids. Following the design of Diffusion Transformers (DiT), the input 3D volume is first tokenized via a \textbf{patchify} operation with a patch size of $2^3$, followed by a linear projection to the model dimension. We utilize absolute position embeddings (APE) to retain spatial information. The core of the network consists of a series of \textbf{Modulated Tri-Attention Blocks}. We incorporate adaptive layer normalization (adaLN) to inject timestep information, modulating the normalized content via scale, shift, and gate parameters.

\paragraph{Sparse Latent Flow Network.} 
For the fine-grained editing of structured latents (Slat), we introduce the SlatFlowModelEditNet. Recognizing the sparsity of high-resolution 3D data, this network adopts a \textbf{hybrid U-Net architecture} that combines the efficiency of sparse convolutions with the global modeling capabilities of transformers. The network features a symmetric encoder-decoder structure with skip connections:
\begin{itemize}
    \item \textbf{Encoder/Decoder:} The encoding path comprises a series of Sparse ResBlocks utilizing sparse convolutions, layer normalizations, and SiLU activations. We employ sparse strided convolutions for downsampling and sparse transposed convolutions for upsampling.
    \item \textbf{Middle Stage:} The bottleneck processing is handled by a stack of \textbf{Sparse Modulated Tri-Attention Blocks}, designed to operate natively on sparse tensors to maximize memory efficiency.
    \item \textbf{Elastic Management:} To handle varying memory loads during training, we integrate an elastic mixing mechanism that dynamically manages gradient flows.
\end{itemize}

\paragraph{Modulated Tri-Attention Mechanism.} 
To effectively integrate multi-modal guidance—maintaining fidelity to the source image while adhering to textual editing instructions—we introduce a specialized Tri-Attention mechanism applied in both networks with domain-specific adaptations.

\noindent\textbf{1) Dense Tri-Attention with KV-Composition.} In the dense structure network, we employ a \textbf{KV-Composer} module to facilitate deep interaction between the visual and textual conditions before the attention operation. The KV-Composer modulates the image context based on the text prompts via an affine transformation supplemented by a low-rank adaptation (LoRA) branch. This injects the semantic editing intent directly into the visual keys and values. Subsequently, a learnable linear \textbf{Mixer} fuses the attention outputs from both modalities, producing a residual update that balances visual preservation and semantic modification.

\noindent\textbf{2) Sparse Tri-Attention.} For the sparse latent network, we adopt a memory-efficient \textbf{Late-Fusion Strategy}. We compute independent sparse cross-attention maps for image and text conditions. Similar to the dense counterpart, a channel-wise mixing layer aggregates these distinct attention flows, generating a unified conditioning signal that guides the flow matching process.

\paragraph{Initialization Details.} 
We follow standard initialization protocols for transformers. Crucially, to ensure training stability, we employ a \textbf{zero-initialization} strategy for the final projection layers of the KV-Composer (LoRA branch), the Mixer, and the adaLN modulation blocks. This ensures that at the initial training stage, the complex multi-modal interaction modules behave as identity functions, progressively learning the editing dynamics.
\subsection{Training Details}
Both the Structured Latent (Slat) and Sparse Structure (SS) models are trained using the Flow Matching framework with an Optimal Transport (OT) path. The objective is to regress the vector field $v_t$ that transports the Gaussian noise distribution to the data distribution, optimized via a standard squared error loss. To bias training towards critical noise levels, we sample time steps $t$ from a Logit-Normal distribution, using parameters $\mu=1.0$ for the Slat model to emphasize structure formation and $\mu=0.0$ for the SS model to ensure balanced diffusion. Conditioning signals are derived from DINOv2 (image) and CLIP (text) encoders, with a $10\%$ random dropout rate applied to enable Classifier-Free Guidance (CFG) during inference.

Optimization is performed using the AdamW optimizer with mixed-precision (FP16) for $1,000,000$ steps at a learning rate of $1 \times 10^{-4}$. We employ adaptive gradient clipping based on historical norms (max 2.0) to stabilize the training dynamics. Crucially, to address the significant memory variance inherent in processing high-resolution sparse grids ($64^3$), we implement an \textit{Elastic Memory Controller} for the Slat model. This mechanism dynamically adjusts the batch workload in real-time to maintain a target GPU memory utilization of $0.75$, ensuring efficient distributed training without out-of-Memory errors.

\subsection{Evaluation Metrics}
\label{subsec:metrics}

To thoroughly evaluate the quality of our generated 3D assets, we employ a comprehensive set of metrics covering geometry accuracy, visual fidelity, semantic alignment, and distribution quality.

\paragraph{Geometry Accuracy.}
We assess the structural quality of the generated meshes using the \textbf{Chamfer Distance (CD)}. Let $S_g$ and $S_r$ be the point clouds sampled from the generated mesh and the reference mesh, respectively. The symmetric Chamfer Distance is defined as:
\begin{equation}
    \text{CD}(S_g, S_r) = \frac{1}{|S_g|} \sum_{x \in S_g} \min_{y \in S_r} \|x - y\|_2^2 + \frac{1}{|S_r|} \sum_{y \in S_r} \min_{x \in S_g} \|y - x\|_2^2,
\end{equation}
where we sample $N=20,480$ points for each set. Lower CD values indicate better geometric reconstruction.
\begin{figure*}
    \centering
    \includegraphics[width=\textwidth]{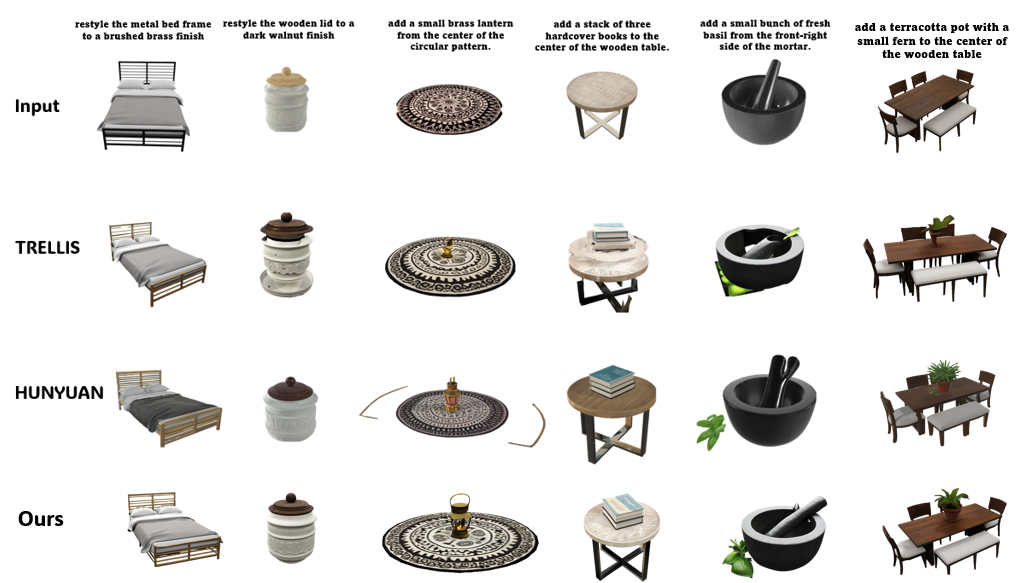}
    \includegraphics[width=\textwidth]{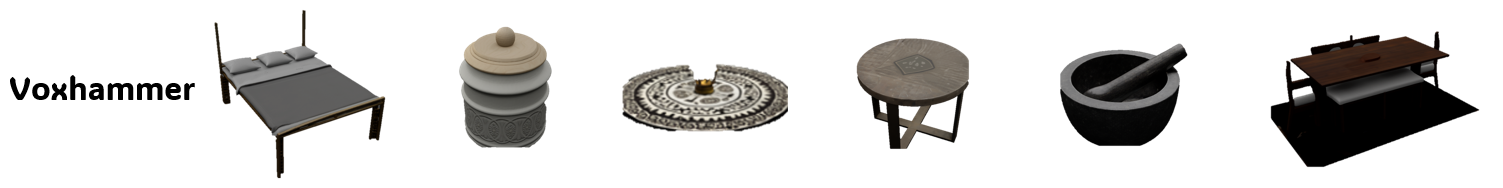}
    \caption{Qualitative comparison with state-of-the-art methods. The first row displays the single input image used for inference. The other rows show the results from baselines and our method. Note that Trellis, VoxHammer, and Ours display renders of the generated 3D assets, whereas Hunyuan shows 2D image editing results. Our method achieves superior 3D consistency and fidelity to the input instruction compared to both 2D-lifting (Trellis) and 2D-editing (Hunyuan) approaches.} 
\end{figure*}
\paragraph{Visual Fidelity and Identity.}
To evaluate appearance preservation and perceptual quality compared to the source reference, we utilize three metrics:
\begin{itemize}
    \item \textbf{SSIM:} The Structural Similarity Index (SSIM) measures the similarity between the rendered view $x$ and the ground truth $y$ based on luminance, contrast, and structure:
    \begin{equation}
        \text{SSIM}(x, y) = \frac{(2\mu_x\mu_y + C_1)(2\sigma_{xy} + C_2)}{(\mu_x^2 + \mu_y^2 + C_1)(\sigma_x^2 + \sigma_y^2 + C_2)},
    \end{equation}
    where $\mu$ and $\sigma^2$ denote the mean and variance, and $\sigma_{xy}$ is the covariance.
    
    \item \textbf{LPIPS:} To capture perceptual similarity closer to human judgment, we compute the Learned Perceptual Image Patch Similarity (LPIPS). It measures the $L_2$ distance between deep features extracted from AlexNet:
    \begin{equation}
        \text{LPIPS}(x, y) = \sum_{l} \frac{1}{H_l W_l} \sum_{h,w} \| w_l \odot (\phi^l(x)_{hw} - \phi^l(y)_{hw}) \|_2^2,
    \end{equation}
    where $\phi^l$ represents the feature map at layer $l$.
    
    \item \textbf{DINO-I:} To quantify high-level structural and identity preservation, we calculate the cosine similarity between DINOv2-Base features:
    \begin{equation}
        \text{DINO-I}(x, y) = \frac{E_{\text{dino}}(x) \cdot E_{\text{dino}}(y)}{\|E_{\text{dino}}(x)\| \|E_{\text{dino}}(y)\|}.
    \end{equation}
    Higher DINO-I scores imply that the edited object retains the core characteristics of the original asset.
\end{itemize}

\paragraph{Semantic Alignment.}
To ensure the edited results strictly follow the text instructions, we calculate the \textbf{CLIP-Score (CLIP-T)}. This metric computes the cosine similarity between the embedding of the generated image $I$ and the text instruction $T$:
\begin{equation}
    \text{CLIP-T}(I, T) = \frac{E_{\text{img}}(I) \cdot E_{\text{txt}}(T)}{\|E_{\text{img}}(I)\| \|E_{\text{txt}}(T)\|},
\end{equation}
using the pre-trained CLIP-ViT-Base-Patch32 model.

\paragraph{Generative Distribution Quality.}
To assess the overall quality and diversity of the generated distribution, as well as temporal consistency for videos, we employ Fréchet-based distances.
\begin{itemize}
    \item \textbf{FID:} The Fréchet Inception Distance (FID) measures the distance between the distribution of real images ($p_r$) and generated images ($p_g$) in the feature space of InceptionV3:
    \begin{equation}
        \text{FID} = \|\mu_r - \mu_g\|_2^2 + \text{Tr}\left(\Sigma_r + \Sigma_g - 2(\Sigma_r \Sigma_g)^{1/2}\right),
    \end{equation}
    where $(\mu, \Sigma)$ represent the mean and covariance of the features.
    
    \item \textbf{FVD:} For video sequences, we utilize the Fréchet Video Distance (FVD). Similar to FID, it computes the distribution distance but uses an I3D network trained on Charades to capture spatiotemporal features, ensuring the temporal coherence of the generated 3D rotations.
\end{itemize}

\section{Comparison with SOTA Methods}
To validate the versatility of BVE in handling both local and global editing tasks, we benchmark it against TRELLIS, HUNYUAN, and VoxHammer\cite{li2025voxhammer}. VoxHammer represents the current state-of-the-art (SOTA) in training-free approaches but relies heavily on edited reference images and explicit mask inputs.
Our results demonstrate that BVE preserves exceptional consistency in non-edited regions, attributed to the robust generative capabilities of the EditFlowTransformer and the regularization provided by our proposed mask loss. In contrast, VoxHammer operates via inversion and attention manipulation within a fixed native latent space. Consequently, it inherently lacks the capacity to perform significant global spatial transformations.
This comparative analysis underscores the superior applicability of our method: BVE achieves SOTA performance in both local and global 3D editing with high fidelity, all while eliminating the need for user-provided masks.
\section{Limitations and Future work}
While our model demonstrates robust capabilities in 3D editing, several limitations remain.
First, regarding the structured latent representation, we employ a two-stage editing pipeline that initially generates the edited sparse structures, followed by the synthesis of the associated local latents. Compared to end-to-end approaches capable of producing complete 3D assets in a single pass, our method may exhibit lower inference efficiency.
Second, the fidelity of our results is heavily contingent upon the capabilities of the underlying Image-to-3D backbone. Consequently, the generated assets often exhibit a strong stylized appearance inherited from the base model. Future improvements will focus on integrating more robust foundation models and enhancing generalization capabilities across diverse editing scenarios. We leave these investigations for future work.

\begin{figure*}[htbp]
    \centering
    \includegraphics[width=\textwidth]{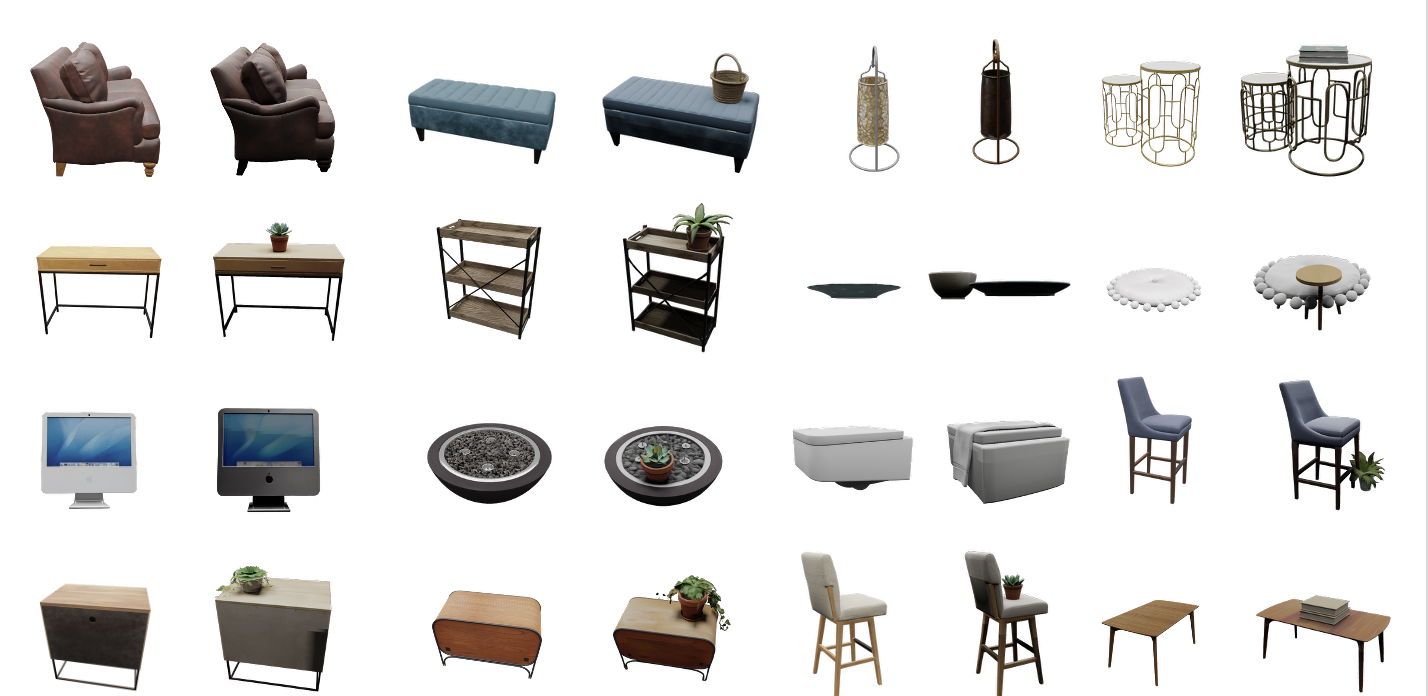}
    \includegraphics[width=\textwidth]{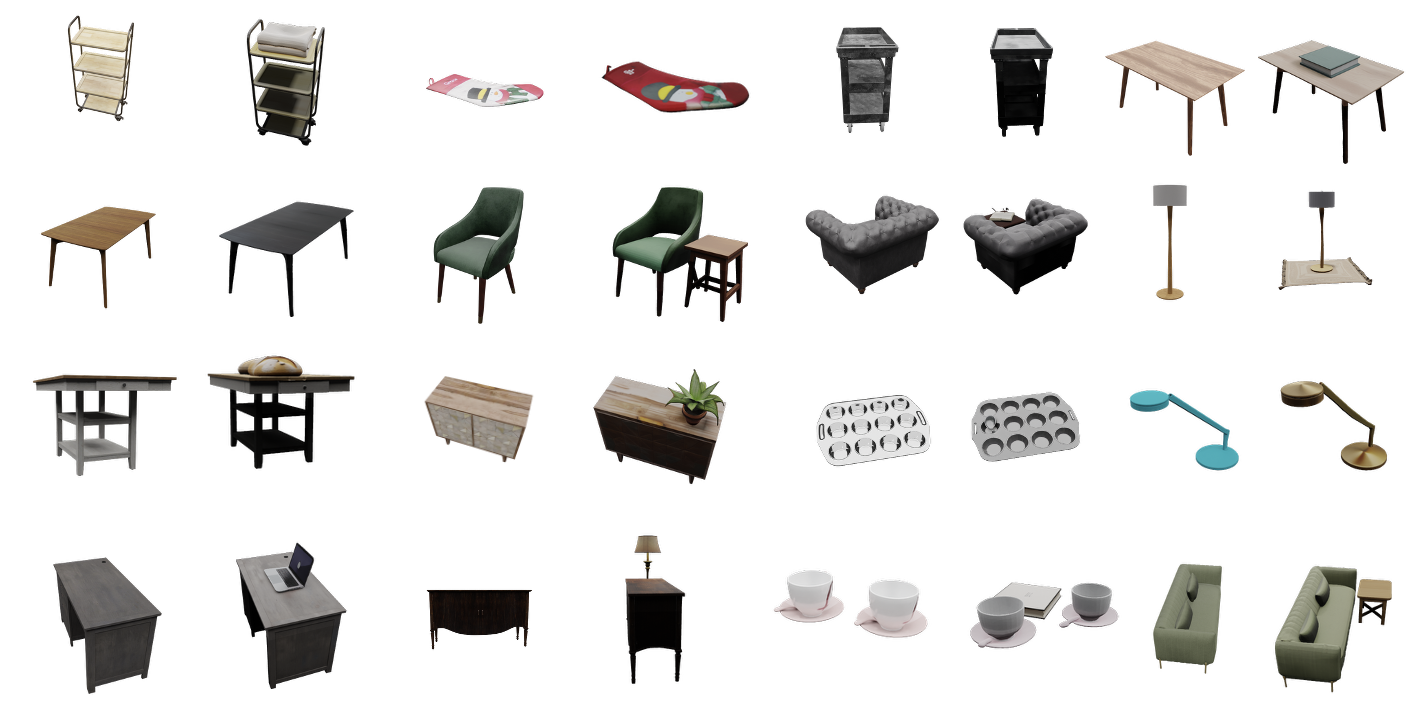}
    \includegraphics[width=\textwidth]{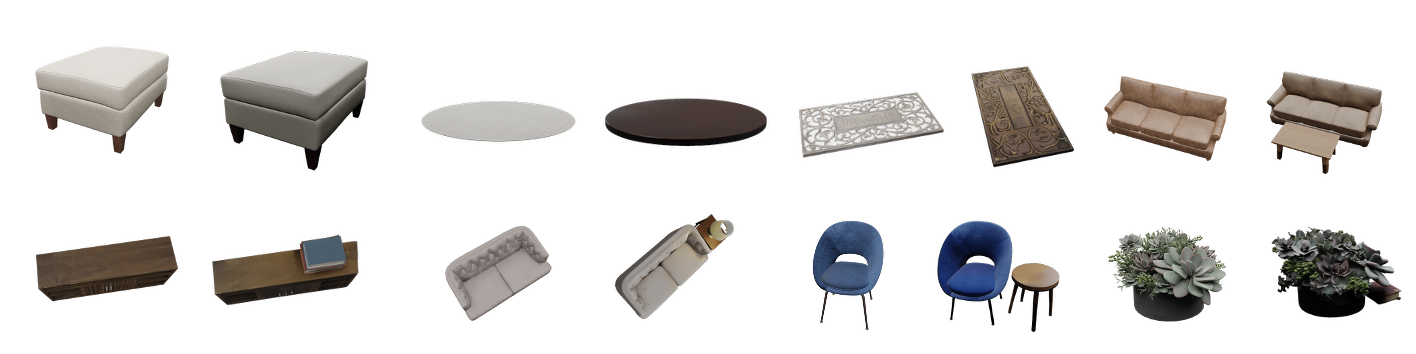}
    \caption{More results generated by with AI Prompts}
    \label{fig:infer_case}
\end{figure*}

\begin{figure*}[htbp]
    \centering
    \includegraphics[width=\textwidth]{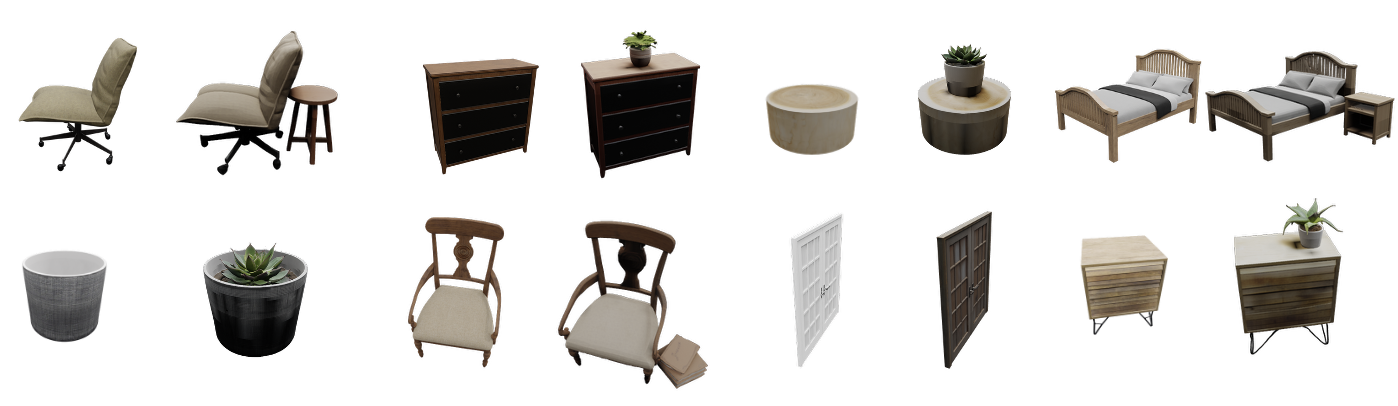}
    \includegraphics[width=\textwidth]{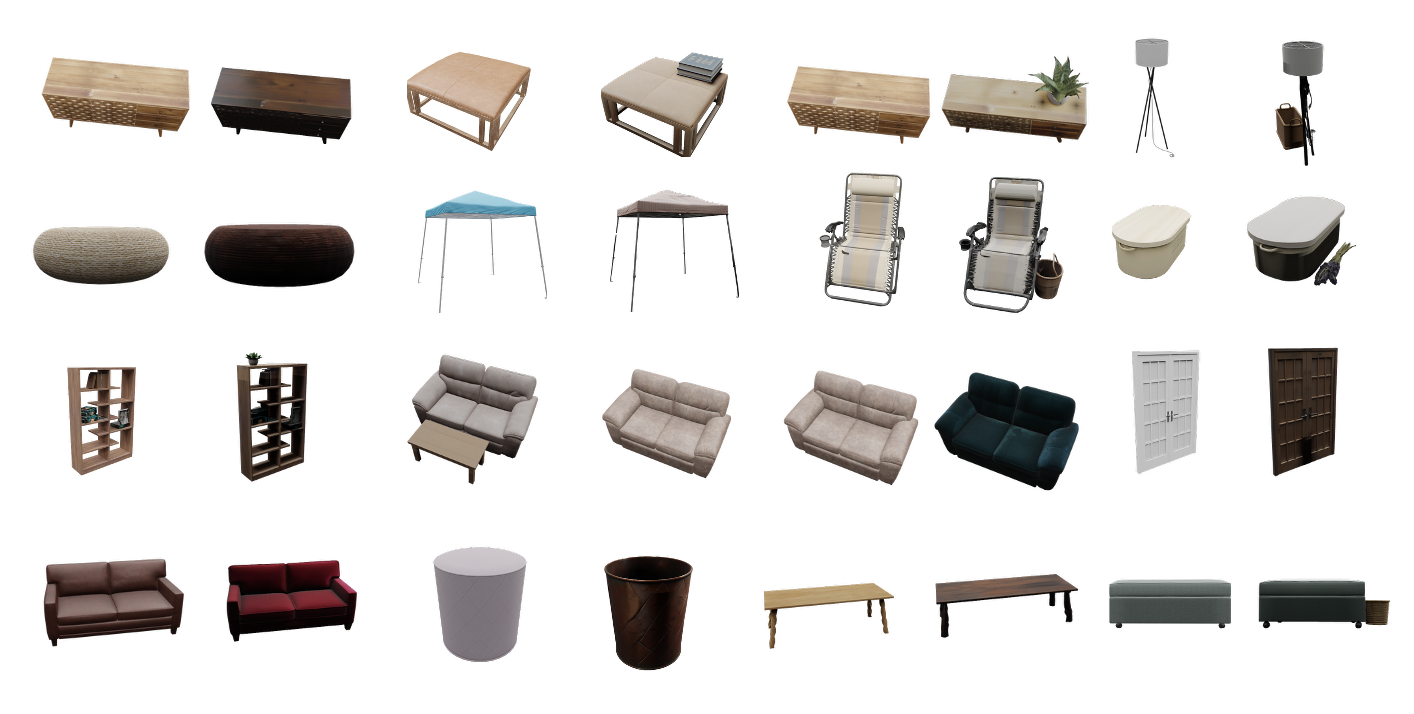}
    \includegraphics[width=\textwidth]{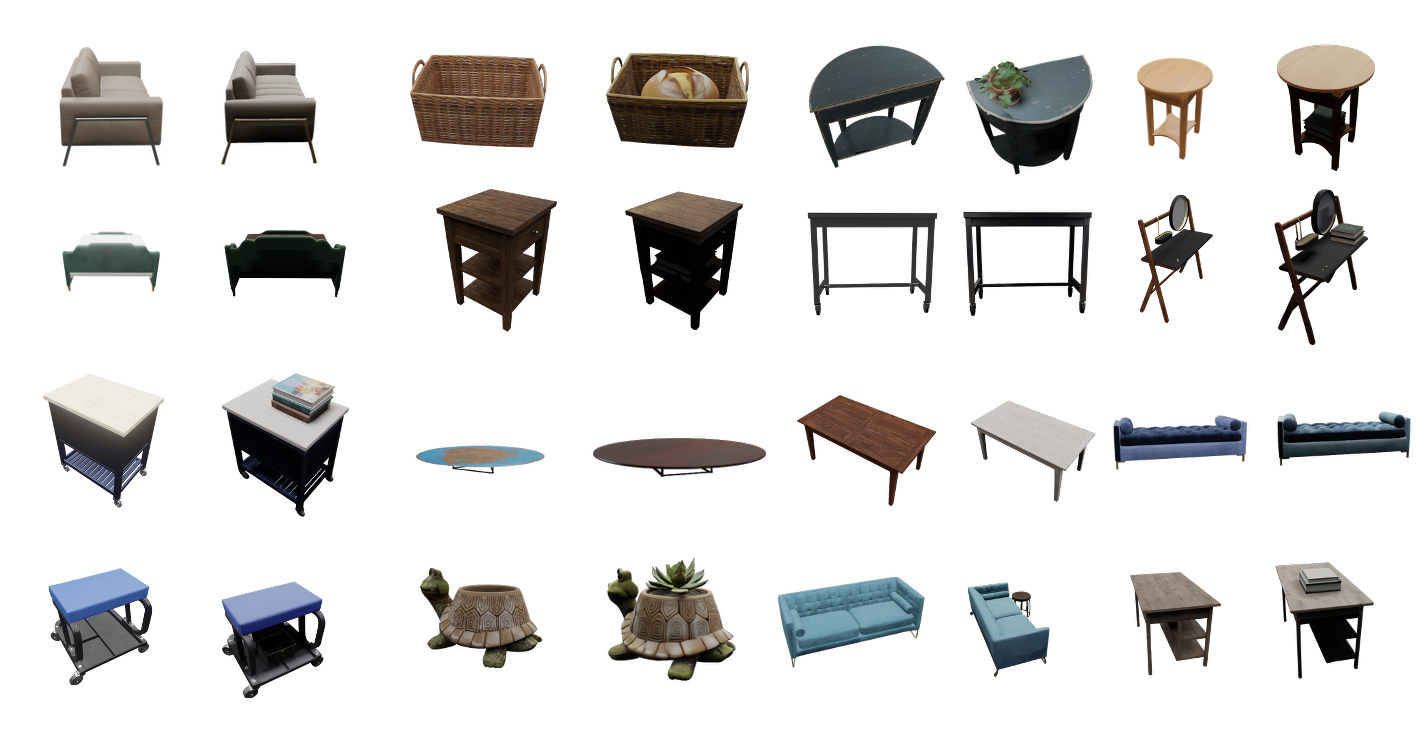}
    \caption{More results generated by with AI Prompts}
    \label{fig:infer_case_2}
\end{figure*}

\end{document}